\documentclass{article} % For LaTeX2e
\usepackage{iclr2027_conference,times}
\ifdefined\pdfsuppressptexinfo\pdfsuppressptexinfo=-1\relax\fi
\ifdefined\pdfinfoomitdate\pdfinfoomitdate=1\relax\fi

\usepackage{amsmath,amsfonts,bm}

\def\eqref#1{equation~\ref{#1}}
\def\1{\bm{1}}

\DeclareMathAlphabet{\mathsfit}{\encodingdefault}{\sfdefault}{m}{sl}
\SetMathAlphabet{\mathsfit}{bold}{\encodingdefault}{\sfdefault}{bx}{n}

\usepackage{hyperref}
\hypersetup{hidelinks}
\usepackage{url}
\usepackage{graphicx}
\usepackage{placeins}
\usepackage{booktabs}
\usepackage{multirow}
\usepackage{array}
\usepackage{amssymb}

\usepackage[font=small]{caption}
\newcolumntype{L}[1]{>{\raggedright\arraybackslash}p{#1}}

\title{Can Motion-Language Models Ground Structure? STRIDE for Evaluating the Evaluators}

\author{Lixing Tan$^{1}$, Qing Xia$^{1}$, Yuting Guo$^{2}$, Shuai Li$^{1}$, Aimin Hao$^{1}$\\
$^{1}$Beihang University\\
$^{2}$Beijing Information Science and Technology University}

\begin{document}

\maketitle

\begin{abstract}
% 【翻译】动作—语言模型通常由动作—语言评测器打分，但这些评测器能在多大程度上
%   将语言结构与动作对应起来，仍不清楚。为此，我们提出 STRIDE（通过时间顺序、
%   镜像反射与动作身份诊断评测结构接地）基准，系统评估评测器识别时间顺序、
%   镜像反射与动作身份的能力。STRIDE 包含 5,869 个三元组，每个三元组由动作、
%   原始描述和扰动描述组成，覆盖短描述与长描述。我们对描述对进行似然配平，
%   以减少纯文本偏置，并估计各评测器在无关动作下的描述偏好，以此为基线衡量
%   匹配动作带来的判别增益。实验揭示了受审计评测器较弱的结构接地能力和严重的
%   镜像敏感性缺陷，而常用评测协议未能暴露这些问题。为理解这些局限为何未在
%   标准测试中被发现，我们进一步考察评测器。我们发现，在现有数据集上，仅凭
%   纯文本先验就能解出朴素的扰动测试；常用检索与分布指标则对镜像真实动作所
%   引入的结构破坏几乎没有响应。这些发现提示了一种自然的干预：结构困难负样本。
%   实验表明，对对比学习的简单修改能大幅改善时间顺序和镜像反射上的表现。
%   基准与代码将公开发布。
Motion-language models are typically scored by motion-language evaluators, 
but how well these evaluators ground language structure remains unclear. Here, we introduce the \textbf{S}tructure grounding
via \textbf{T}emporal-order, \textbf{R}eflection, and \textbf{I}dentity
\textbf{D}iagnostic \textbf{E}valuation (STRIDE) benchmark to systematically evaluate the ability of evaluators to
track temporal order, mirror reflection, and action identity. STRIDE comprises $5{,}869$ triples, each consisting of a
motion, its original caption, and a perturbed caption, spanning both short
and long descriptions. We likelihood-balance caption pairs to reduce text-only
bias and estimate each evaluator's caption preference under unrelated motions
to measure the discrimination gain from matched motions relative to this
baseline. Our experiments reveal weak structural grounding and severe
deficits in mirror sensitivity among the audited evaluators, which
commonly used evaluation protocols fail to expose. To understand why
these limitations go undetected in standard tests, we examine the
evaluators more closely. We find that text-only priors alone can solve
naive perturbation tests on existing datasets, while common retrieval
and distributional metrics barely respond to structural corruption
introduced by mirroring ground-truth motions. These findings suggest
a natural intervention: structural hard negatives. Our experiments show
that a simple modification to contrastive learning substantially improves
performance on temporal order and mirror reflection. The benchmark and
code will be released.
\end{abstract}

% ======================================================================
% 【翻译】引言。
\section{Introduction}
\begin{figure}[htb]
\centering
\includegraphics[width=\linewidth]{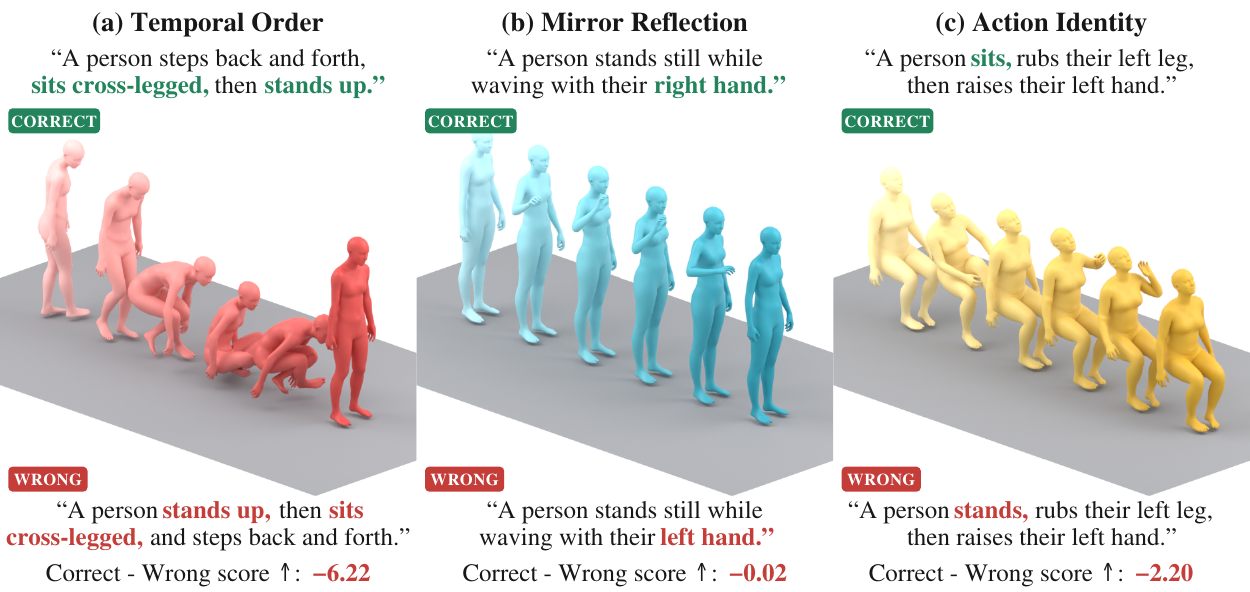}
% 【翻译】HumanML3D 评测器的结构接地失败。每个面板固定一个真实动作，并沿时间
%   顺序、镜像反射或动作身份中的一条轴扰动描述。三个例子中，正确描述得分减去
%   错误描述得分的差值均为负，表明评测器将错误描述排在正确描述之前。
\caption{Structural grounding failures of the HumanML3D evaluator.
Each panel holds a fixed real motion and perturbs its caption along one axis: temporal order, mirror reflection, or action identity. In all three examples, the correct$-$wrong score is negative, indicating the evaluator ranks the incorrect caption above the correct one.}
\label{fig:running}
\end{figure}

% ======================================================================
% 【翻译】动作生成与理解在具身智能、计算机动画和世界模型仿真等领域有着重要应用。
%   当前评估通常利用预训练的动作—语言评测器提取嵌入，计算 R-Precision、
%   Fréchet Inception Distance（FID）和匹配距离（MM-Dist）等指标。一个广泛
%   使用的例子是 HumanML3D 引入的对比式文本—动作嵌入评测器。然而，这些指标
%   上的表现是否真的说明模型理解了动作与文本的对应关系？问题在于，如果评测器
%   本身无法区分结构不同的动作或描述，基于其嵌入空间计算的指标也难以揭示相应
%   错误。图 fig:running 展示了三个具体例子：固定每个真实动作，仅改变其描述
%   中的时间顺序、镜像反射或动作身份，再用 HumanML3D 评测器为得到的配对打分。
%   令人意外的是，评测器在三个例子中均偏好扰动后的错误描述。这些观察揭示了
%   HumanML3D 评测器区分细粒度结构差异的失败，并引出进一步的问题：其他评测器
%   是否也存在类似缺陷？
Motion generation and understanding have important applications in
embodied intelligence, computer animation, and world-model
simulation~\citep{tevet2023mdm, zhang2022motiondiffuse, chen2023mld,
zhang2023remodiffuse, zhang2023t2mgpt, guo2022tm2t,
guo2024momask}.
Current evaluation commonly uses pretrained motion-language evaluators to
extract embeddings for computing metrics such as R-Precision, Fr\'echet
Inception Distance (FID)~\citep{heusel2017fid}, and Motion Matching Distance
(MM-Dist). A widely used example is the contrastive text-motion embedding
evaluator introduced with HumanML3D~\citep{guo2022humanml3d}.
But does performance on these metrics really indicate that a model
understands the correspondence between motion and text? The concern is
that, if an evaluator itself cannot distinguish structurally different
motions or captions, metrics computed in its embedding space will also
struggle to expose the corresponding errors.
Figure~\ref{fig:running} presents three concrete examples: we fix each real
motion, change only temporal order, mirror reflection, or action identity
in its caption, and score the resulting pairs with the HumanML3D evaluator.
Surprisingly, the evaluator prefers the perturbed, incorrect caption in
all three examples. These observations expose failures of the HumanML3D
evaluator to distinguish fine-grained structural differences and raise a
further question: do other evaluators exhibit similar shortcomings?

% 【翻译】要回答这一问题，首先需要审视常规训练与评测是否充分要求模型辨别结构差异。许多
%   动作—语言匹配器通过对比学习区分配对样本与非配对样本，检索评测则检验
%   它们能否找回正确配对。例如，给定一段抬左手的动作，如果候选描述分别为
%   “抬左手”“行走”和“坐下”，评测器只需识别“抬手”这一整体动作就可能选对，
%   而无须辨别左右。相比之下，要根据实际动作可靠地区分“抬左手”和“抬右手”，评测器需要
%   将描述中的左右关系与实际动作对应起来。因此，当训练和评测中的非匹配
%   候选主要在整体动作内容上不同，良好的匹配表现并不一定意味着模型能够辨别
%   细粒度结构差异。然而，即使候选描述仅在目标结构上不同，选对答案也不一定
%   说明评测器使用了动作信息：它仍可能依赖语言流畅度或描述偏好。为此，我们提出
%   STRIDE，一个覆盖时间顺序、镜像反射和动作身份的诊断评测基准，以无关动作
%   下的描述偏好为基线，检验匹配动作是否改善评测器对结构差异的判别。
%   表 1 对比了 STRIDE 与现有评测协议在结构覆盖和偏好控制方面的差异。
Answering this question first requires examining whether standard training
and evaluation sufficiently demand structural discrimination.
Many motion-language matchers use contrastive training
to distinguish paired from unpaired samples, while retrieval evaluates
whether they recover the correct
pair~\citep{petrovich2023tmr,chen2024clam,guo2025snapmogen}.
For example, given a motion of someone raising their left hand, an
evaluator choosing among ``raising the left hand,'' ``walking,'' and
``sitting down'' may succeed by recognizing hand raising without
distinguishing left from right. In contrast, reliably distinguishing
``raising the left hand'' from ``raising the right hand'' based on the
observed motion requires grounding the left--right distinction in that
motion. Thus, when nonmatching candidates
in training and evaluation differ mainly in overall action content,
successful matching need not imply fine-grained structural
discrimination~\citep{yuksekgonul2023aro}.
However, even when candidate captions differ only in the target structure,
a correct choice does not necessarily indicate that the evaluator used
motion information: it may still rely on fluency or caption
preference~\citep{hsieh2023sugarcrepe}.
We therefore introduce STRIDE, a diagnostic benchmark covering temporal
order, mirror reflection, and action identity, which uses caption preference
under unrelated motions as a baseline to assess whether matched motions
improve evaluators' discrimination of structural differences.
Table~\ref{tab:protocols} compares STRIDE with existing evaluation protocols
in structural coverage and preference control.

\begin{table}[htb]
% 【翻译】与现有动作—语言评测协议的比较。STRIDE 将时间顺序、镜像反射和动作
%   身份的受控测试与文本侧偏置检查、评测器特定的描述偏好控制相结合，同时覆盖短描述与长描述。
\caption{Comparison with existing motion-language evaluation protocols.
STRIDE combines controlled tests of temporal order, mirror reflection,
and action identity with text-only bias checks and evaluator-specific
caption-preference controls across both short and long descriptions.}
\label{tab:protocols}
\centering
\scriptsize
\setlength{\tabcolsep}{1.0pt}
\renewcommand{\arraystretch}{1.05}
\begin{tabular*}{\linewidth}{@{\extracolsep{\fill}}L{0.26\linewidth}cccccccc@{}}
\toprule
\textbf{Evaluation Setting} &
\shortstack{\textbf{Content}\\\textbf{Matching}} &
\shortstack{\textbf{Temporal}\\\textbf{Order}} &
\shortstack{\textbf{Mirror}\\\textbf{Reflection}} &
\shortstack{\textbf{Action}\\\textbf{Identity}} &
\shortstack{\textbf{Matched}\\\textbf{Pairs}} &
\shortstack{\textbf{Prior}\\\textbf{Controls}} &
\shortstack{\textbf{Short}\\\textbf{Captions}} &
\shortstack{\textbf{Long}\\\textbf{Captions}} \\
\midrule
HumanML3D suite~\citep{guo2022humanml3d} &
\checkmark & -- & -- & -- & -- & -- & \checkmark & -- \\
TMR protocols~\citep{petrovich2023tmr} &
\checkmark & -- & -- & -- & -- & -- & \checkmark & -- \\
CLaM~\citep{chen2024clam} &
\checkmark & -- & -- & -- & -- & -- & \checkmark & -- \\
MoBERT~\citep{voas2023metric} &
-- & -- & -- & -- & -- & -- & \checkmark & -- \\
ChronRet (CAR)~\citep{fujiwara2024car} &
\checkmark & \checkmark & -- & -- & \checkmark & -- & \checkmark & -- \\
SnapMoGen suite~\citep{guo2025snapmogen} &
\checkmark & -- & -- & -- & -- & -- & -- & \checkmark \\
\midrule
\textbf{STRIDE (ours)} &
\checkmark & \checkmark & \checkmark & \checkmark & \checkmark & \checkmark &
\checkmark & \checkmark \\
\bottomrule
\end{tabular*}

\end{table}

% 【翻译】我们的贡献有三项：
%   (1) 我们提出 STRIDE，用于细粒度评估动作—语言评测器的结构接地能力，
%   围绕时间顺序、镜像反射和动作身份设计受控任务，覆盖短描述与长描述。
%   我们结合描述偏好控制与随机内容参照，评估各条结构轴上的能力。
%   (2) 我们系统评估六个已发布的动作—语言评测器，揭示其整体内容匹配与
%   细粒度结构区分之间的能力缺口。在控制无关动作下的描述偏好后，所有受测
%   评测器在三条结构轴上的接地表现均低于各自的随机内容参照，镜像反射是
%   多数评测器共同的薄弱环节。
%   (3) 我们进一步分析，当负样本主要在整体内容上存在差异时，对比训练为何
%   可能缺少学习细粒度结构差异的要求。基于这一分析，我们引入结构感知的
%   困难负样本机制。实验表明，该机制能够改善受测评测器的时间顺序与镜像
%   反射接地能力，收益随评测器和结构轴而异。
Our contributions are threefold:
\begin{itemize}
\item We introduce STRIDE, a benchmark for fine-grained evaluation of
structural grounding in motion-language evaluators, with controlled tasks
covering temporal order, mirror reflection, and action identity across
short and long captions. We combine caption-preference controls with a
Random Content reference to assess capabilities along each structural axis.
\item We systematically evaluate six released motion-language evaluators
and reveal a gap between overall content matching and fine-grained structural
discrimination. After controlling for caption preference under unrelated
motions, all tested evaluators achieve lower grounding scores on each
structural axis than on their Random Content reference, with Mirror
Reflection a recurring weakness for most evaluators.
\item We analyze why contrastive training may not require fine-grained
structural discrimination when negatives differ mainly in overall content. Based on this analysis, we introduce a
structure-aware hard-negative scheme. Experiments show that this scheme
can improve temporal-order and mirror-reflection grounding in the tested
evaluators, with gains varying across evaluators and structural axes.
\end{itemize}

% ======================================================================
% 【翻译】STRIDE 基准：评测器何时会忽略结构？
\section{The STRIDE benchmark: When do evaluators miss structure?}
\label{sec:benchmark}
% ======================================================================

\begin{figure}[htb]
\centering
\includegraphics[width=\linewidth]{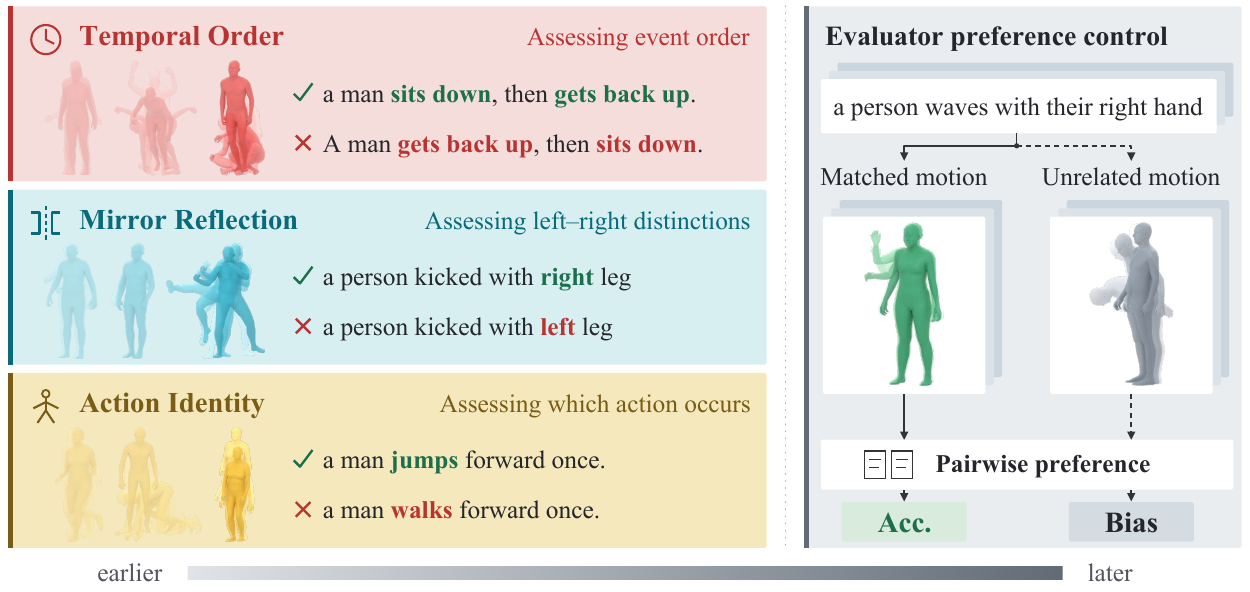}
% 【翻译】STRIDE 检验三种结构区分，并在无关动作控制下测量各评测器的描述偏好。
%   左：每条深色动作序列与其原始和编辑描述配对；高亮短语标出结构改动，
%   由浅到深的着色编码相对时间。右：绿色动作序列与描述匹配，灰色序列与描述
%   无关，用于估计模型偏好。每个评测器分别在匹配动作和无关动作条件下评估
%   动作—描述匹配，得到 acc 和 bias。
\caption{STRIDE probes three structural distinctions and measures each
evaluator's caption preference under an unrelated-motion control.
Left: each dark motion sequence is paired with its original and edited
captions; highlighted phrases mark the structural change, and
lighter-to-darker shading encodes relative time. Right: green motion
sequences match the caption, whereas grey sequences are unrelated to it
and serve to estimate model preference. Each evaluator assesses
motion-caption matching under the matched- and unrelated-motion
conditions to obtain acc and bias, respectively.}
\label{fig:construction}
\end{figure}

% 【翻译】如图 fig:construction 所示，我们提出基于 HumanML3D 和 SnapMoGen
%   构建的 STRIDE 基准，检验动作—语言评测器能否识别结构差异。本节先定义时间顺序、镜像反射和动作身份
%   三条扰动轴（第 2.1 节），再介绍基准的构建与质量检查（第 2.2 节）。
%   随后，我们引入无关动作对照，衡量匹配动作相对于描述偏好基线带来的
%   判别增益（第 2.3 节），并比较已发布评测器在各条轴与数据集上的表现
%   （第 2.4 节）。
As shown in Figure~\ref{fig:construction}, we introduce STRIDE, a benchmark built from HumanML3D
and SnapMoGen to validate whether motion-language
evaluators recognize structural differences.
We first define the Temporal Order, Mirror Reflection, and Action Identity
perturbations (Section~\ref{sec:structural-tests}), then describe benchmark
construction and quality checks (Section~\ref{sec:benchmark-construction}).
Next, we introduce unrelated-motion controls to measure the discrimination
gain from matched motions over the caption-preference baseline
(Section~\ref{sec:grounding-protocol}). Finally, we compare released
evaluators across structural axes and datasets (Section~\ref{sec:results}).

% 【翻译】结构扰动。
\subsection{Structural perturbations}
\label{sec:structural-tests}
% 【翻译】我们通过修改动作描述中的一个目标属性来构造结构扰动，同时保持
%   对应动作及描述中的其他信息不变。STRIDE 考察三类变化：时间顺序改变多个
%   动作事件的先后关系；镜像反射交换描述中的左右关系；动作身份将一个动作
%   替换为另一个视觉上可区分的动作。这些测试要求评测器区分具体的结构差异。为对照结构判别与整体内容匹配，我们把匹配描述换成语料中另一段动作的描述，
%   作为随机内容匹配参照。每个 STRIDE 条目是一个三元组 (m, c, c′)：一段真实动作 m、它的真值
%   描述 c、以及只改变一个目标属性的扰动句 c′（图 2）。评测器以 m 为输入，
%   在 c 与 c′ 之间选择。
We construct structural perturbations by changing one target property of
a motion caption while keeping the motion and the other caption information
fixed. STRIDE tests three types of changes: \textbf{Temporal Order}
changes the order of events in a multi-action caption; \textbf{Mirror
Reflection} exchanges left and right in the description; and \textbf{Action
Identity} replaces one action with a visually distinct alternative.
These tests ask whether evaluators distinguish specific structural
differences. To compare structural discrimination with overall content matching, we replace the matched caption with the caption of a
different corpus motion as a reference of \textbf{Random Content} matching. Each STRIDE item is a triplet $i=(m, c, c')$: a real motion $m$,
its ground-truth caption $c$, and a single-property perturbation $c'$. Given $m$, the evaluator chooses between
$c$ and $c'$.

% 【翻译】时间顺序编辑保留原有动作、主体与场景。镜像编辑交换所有方向性描述left/right，
%   保持描述流畅，并对应物理上有效的镜像动作。
%   动作身份编辑使用预先定义的动作词对（如 runs→walks），保留词形以及
%   主语、宾语等句子成分。
Temporal edits preserve the original actions, actor, and scene. Mirror
edits replace every directional description (e.g., \emph{left}$\to$\emph{right}), preserving fluency and
corresponding to a physically valid mirrored motion. Action Identity
edits use predefined verb pairs (e.g., \emph{runs}$\to$\emph{walks})
while preserving grammatical form and arguments, such as the subject and
objects.

% 【翻译】基准构建。
\subsection{Benchmark construction}
\label{sec:benchmark-construction}
% 【翻译】基准使用 HumanML3D 与 SnapMoGen 的官方测试划分，分别覆盖短片段
%   描述与长描述。
The benchmark uses the official HumanML3D and SnapMoGen test
splits, which provide short
clip-level captions and long-form descriptions, respectively.

% 【翻译】我们从每条真值描述出发构造扰动。时间顺序将多动作描述中的事件
%   顺序反转，同时保留所有动作；镜像反射交换全部 left/right 词；动作身份
%   按预先定义的动作词对替换一个动词，并保留其语法形式。动作身份候选另经
%   独立的纯文本检查，不呈现动作。具体实现见附录 app:datasheet。
We construct perturbations from each ground-truth caption. For Temporal
Order, we reverse the event order in multi-action descriptions while
retaining every action. For Mirror Reflection, we swap all occurrences of
\emph{left} and \emph{right}. For Action Identity, we replace one verb
using the predefined action pairs while preserving its grammatical form.
Action Identity candidates also undergo an independent text-only check,
with no motion shown. Implementation details are provided in
Appendix~\ref{app:datasheet}.

% 【翻译】扰动后的描述可能不如原句流畅，评测器因此可能仅凭文本流畅度选对描述。
%   例如，时间顺序扰动将
%   “a person is waving their right arm up and down and then side to side”变为
%   “a person side to side then is waving their right arm up and down”。
%   后一句开头缺少谓语，使评测器无需查看动作即可利用语法线索作答。
%   我们用似然筛选限制这种差异。令 ℓ̄_L(c) 为描述 c 在 GPT-2 Large 下的平均
%   token 对数似然，定义
%   描述先验差Δℓℓ(c,c′)=ℓ̄_L(c)−ℓ̄_L(c′)。正值表示仅凭文本即偏向原句。结构
%   句对仅当绝对差不超过每 token 0.15 nats 时保留。
An edited caption may be less fluent than the original, allowing an
evaluator to choose correctly from text fluency. For example, a Temporal Order perturbation changes ``a person is waving their
right arm up and down and then side to side'' to ``a person side to side
then is waving their right arm up and down.'' The missing verb in the
opening clause provides a grammatical cue without requiring the evaluator
to inspect the motion. We limit this difference
with a likelihood filter. With
$\bar\ell_L(c)$ the mean token log-likelihood of caption $c$ under GPT-2
Large~\citep{radford2019gpt2}, the caption-prior gap is defined as follows:
\begin{equation}
    \Delta\ell\ell(c,c')=\bar\ell_L(c)-\bar\ell_L(c')
\end{equation}
where positive values
favor the original caption from text alone; structural pairs are kept when
$|\Delta\ell\ell|\le 0.15$ nats per token.

\begin{figure}[!htbp]
\centering
\includegraphics[width=\linewidth]{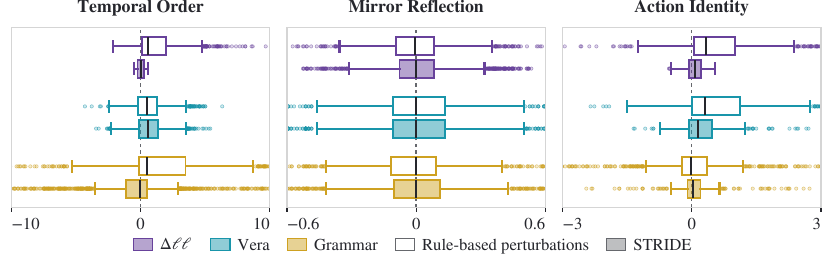}
% 【翻译】规则式扰动与 STRIDE 的纯文本分差比较。颜色区分 Δℓℓ、Vera 与 Grammar；
%   空心箱和实心箱分别表示规则式扰动与 STRIDE。分差为原句得分减去扰动句得分，
%   并按各打分器的规则式扰动结构样本四分位距（IQR）缩放；零表示两句得分相同。
%   两个样本池并非逐对前后对照。
\caption{Comparison of text-only score gaps between rule-based perturbations and STRIDE.
Colors distinguish $\Delta\ell\ell$, Vera, and Grammar; open and filled
boxes represent rule-based perturbations and STRIDE, respectively. Gaps are original
minus perturbed scores, scaled by each scorer's interquartile range (IQR)
on the rule-based structural pairs; zero indicates equal scores. The two pools are
not matched before/after samples.}
\label{fig:bias-necessity}
\end{figure}

% 【翻译】为检查构造后的描述对是否仍包含可利用的纯文本线索，我们使用 GPT-2、
%   Vera 和 Grammar，分别检查语言似然、常识合理性和语法可接受性的分差，
%   并将 STRIDE 与规则式扰动进行比较。如图 fig:bias-necessity 所示，STRIDE 在
%   时间顺序和动作身份轴上的分差更集中于零附近，表明构造数据中的部分纯文本
%   偏好有所减少。
To check whether the constructed caption pairs still contain exploitable
text-only cues, we use GPT-2, Vera, and Grammar to examine gaps in language
likelihood, commonsense plausibility, and grammatical acceptability,
respectively, and compare STRIDE with rule-based perturbations.
As shown in Figure~\ref{fig:bias-necessity}, STRIDE's score gaps on Temporal
Order and Action Identity are more concentrated near zero, indicating
reduced text-only preferences in some aspects of the constructed data.

% 【翻译】最后，人工审计通过检查描述流畅性、是否只改变一个属性，以及能否从动作中
%   区分该变化，评估 STRIDE 的条目质量。发布的基准含 4,269 个结构句对和
%   1,600 个随机内容参照：共 5,869 个三元组，覆盖 2,598 段动作。
Finally, a human audit assesses item quality in STRIDE by checking caption
fluency, whether only one property has changed, and whether the change
can be distinguished from the motion. The released benchmark contains $4{,}269$ structural pairs
plus $1{,}600$ Random Content references: $5{,}869$ triples over $2{,}598$ motions.

\begin{figure}[!htbp]
\centering
\includegraphics[width=\linewidth]{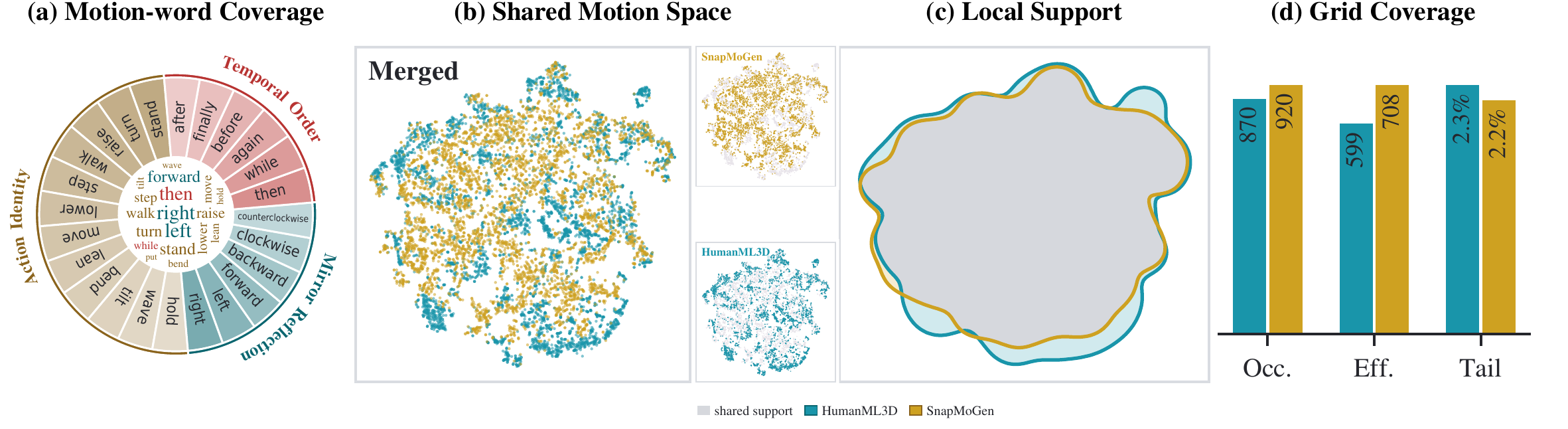}
% 【翻译】词汇与动作覆盖。(a) 各轴高频词。(b) 共享特征的 t-SNE 投影。
%   (c) 平滑的语料覆盖区域。(d) 55×55 网格上的占用格数（Occ.）、有效格数
%   （Eff.）和长尾占比（Tail）；柱高按指标归一化，标注保留未经缩放的数值。
\caption{Vocabulary and motion coverage. (a) Frequent words by axis.
(b) Shared-feature t-SNE projection. (c) Smoothed corpus footprints.
(d) Occupied cells (Occ.), effective cells (Eff.), and long-tail share
(Tail) on a $55\times55$ grid; bars are normalized per metric, with
unscaled labels.}
\label{fig:coverage}
\end{figure}

% 【翻译】图 fig:coverage 展示各轴的词频分布，以及每个源语料各采样 5,000 段动作的
%   覆盖统计。两套语料在共享运动学投影中大幅重叠，SnapMoGen 的网格占用分布
%   更均匀；统计方法见附录 app:coverage。
Figure~\ref{fig:coverage} shows word frequencies by axis and motion-coverage
statistics for $5{,}000$ clips sampled from each source corpus. The corpora
overlap broadly in the shared kinematic projection, with SnapMoGen more
evenly spread across the grid. Appendix~\ref{app:coverage} details the analysis.

% 【翻译】引入无关动作对照的评估。
\subsection{Evaluation with unrelated-motion controls}
\label{sec:grounding-protocol}
% 【翻译】受控描述对将测试聚焦于目标差异，但选对描述本身仍不足以说明匹配
%   动作提供了帮助。因此，我们进一步追问：与接收无关动作相比，评测器在
%   接收匹配动作时能否更可靠地区分同一对描述？
Controlled caption pairs focus the test on the targeted distinction, but a
correct choice alone does not establish that the matched motion helped.
We therefore ask a further question: does the evaluator distinguish the
same captions more reliably with the matched motion than with an unrelated
motion?

% 【翻译】原始准确率可能同时反映动作判别能力和描述偏好：如果原句更自然，
%   评测器即使不利用动作，也可能取得高于随机水平的准确率。为衡量观测动作带来的实际增益，
%   我们保持描述不变，分别使用匹配动作和无关动作进行评估。
%   对条目 i，令 d_i 和 b_i 分别为匹配动作和无关动作条件下原句得分减去扰动句得分的差值。
%   两种条件独立计分：分差为正、零、负时，分别记为 1、0.5、0。
%   对每个评测器，在每个数据集的每条扰动轴内分别取平均，得到匹配动作下的准确率 acc，
%   以及无关动作下偏好原句的比例 bias。
%   我们用 acc−bias 衡量匹配动作带来的准确率增益，再除以从该基线到完全正确的
%   剩余提升空间 1−bias，得到归一化增益 Δ：
Raw accuracy can reflect both motion discrimination and caption preference:
if original captions are more natural, an evaluator can score above
chance without using the motion. To measure the actual gain from the observed
motion, we keep the captions fixed and evaluate them with matched
and unrelated motions. For item $i$, let $d_i$ and $b_i$ denote the original-caption
score minus the perturbed-caption score under the matched motion and the
unrelated motion $\tilde m_i$, respectively. Each condition is scored
separately: positive, zero, and negative margins receive $1$, $0.5$, and
$0$, respectively. For each evaluator, we average these values separately
within each corpus and axis to obtain matched-motion accuracy
$\mathrm{acc}$ and the original-caption preference rate under unrelated
motions, $\mathrm{bias}$. We measure the accuracy gain from matched
motions as $\mathrm{acc}-\mathrm{bias}$ and divide it by the remaining
room for improvement, $1-\mathrm{bias}$, to obtain the normalized gain:
\begin{equation}
    \Delta
  =\frac{\mathrm{acc}-\mathrm{bias}}{1-\mathrm{bias}}
\end{equation}
% 【翻译】其中，Δ 表示识别匹配动作与描述之间结构差异的能力，越接近 1 越好。
where $\Delta$ denotes the ability to recognize the structural gap between matched motions and captions. Values of $\Delta$ closer to $1$ indicate better performance.

% 【翻译】Δ 衡量总体准确率的提升，但评分的改善不一定会改变最终选择。
%   为捕捉这类改善，我们提出接地扰动识别率（GPR）。GPR 衡量有多大比例的样本
%   在匹配动作下的原句减扰动句分差，大于无关动作下的同一分差，计算如下：
%   GPR=Pr[d_i>b_i]+0.5Pr[d_i=b_i]。
%   GPR 越接近 100%，说明匹配动作越能在更多样本上将评分推向正确的结构判断。
While $\Delta$ measures the overall accuracy gain, an improvement in scores
need not change the final choice. To capture such improvements, we propose
\emph{Grounded Perturbation Recognition} (GPR). GPR measures the proportion of items
whose original-minus-perturbed caption margin is larger under the matched
motion than under the unrelated motion. GPR is calculated as follows:
\begin{equation}
    \mathrm{GPR}=\Pr[d_i>b_i]+\tfrac{1}{2}\Pr[d_i=b_i]
\end{equation}
where a GPR closer to $100\%$ indicates that matched motions push scores toward
the correct structural interpretation on more items.

\begin{figure}[!htb]
\centering
\includegraphics[width=\linewidth]{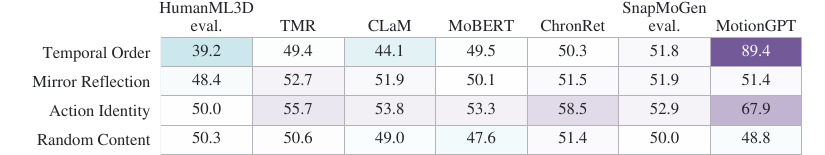}
% 【翻译】不同模型在无关动作下的描述偏好比较。每格报告模型偏好原句而非
%   扰动句的比例（bias，%，平局记半分）。白色表示 50%；紫色和青色分别表示
%   偏向原句和扰动句。
\caption{Comparison of caption preferences across models under unrelated motions.
Cells report the rate of preferring the original over the perturbed caption
(bias, \%; ties count as half). White marks $50\%$; purple and teal indicate
preferences for the original and perturbed captions, respectively.}
\label{fig:evaluator-priors}
\end{figure}

% 【翻译】为检验上述偏好对照的必要性，图 fig:evaluator-priors 展示各模型在
%   无关动作下的原句偏好率 bias。
%   即使经过数据侧筛选，这些偏好率仍随模型和扰动轴变化，说明受测模型在该对照
%   条件下仍存在一定描述偏好。
To examine the need for these controls, Figure~\ref{fig:evaluator-priors}
reports each model's original-caption preference rate (bias) under
unrelated motions. Even after data-side filtering, these rates vary across
models and perturbation axes, indicating that the evaluated models still
exhibit some caption preferences under this control.

% 【翻译】评估结果。
\subsection{Evaluation results}
\label{sec:results}
% 【翻译】以上述无关动作偏好作为基线，我们进一步考察：评测器从匹配动作中
%   获得的结构区分收益是否仍低于整体内容匹配，以及这一差距主要出现在哪些
%   评测器和结构轴上。主要审计对象是已发布评测器；生成模型和视觉语言模型
%   的适配接口作为补充测试单独讨论。
Using these unrelated-motion preferences as baselines, we next examine
whether matched-motion gains in structural discrimination remain below
those in overall content matching, and which evaluators and axes show
the largest gaps. Our primary audit concerns released evaluators;
adapted generative and vision-language interfaces are examined separately
as supplementary tests.

% 【翻译】表 2 审计六个已发布评测器：HumanML3D 块含 HumanML3D 评测器、TMR、
%   CLaM、MoBERT 与经过时序训练的 ChronRet，均使用各自的原生表示；
%   SnapMoGen 块含原生 SnapMoGen 评测器与五个 HumanML3D 训练评测器。表中逐轴报告 acc、bias 与 Δ，并给出宏平均 GPR。
Table~\ref{tab:main} audits six released evaluators: the HumanML3D block
holds the HumanML3D evaluator, TMR, CLaM, MoBERT, and the
chronology-trained ChronRet on their
native representation, and the SnapMoGen block holds the native SnapMoGen evaluator plus the five HumanML3D-trained evaluators. Table~\ref{tab:main} reports acc, bias, and $\Delta$ per axis, plus macro GPR.

\begin{table}[ht]
% 【翻译】现有动作—语言评测器在 STRIDE 上的性能比较。我们报告匹配动作下的
%   准确率（acc）、无关动作下的原句偏好率（bias）及归一化接地增益 Δ；
%   acc 和 bias 以小数比例表示。Macro GPR（%）为三条结构轴的 GPR 均值，
%   Random Content 作为内容匹配参照。加粗和下划线分别标记每个数据集分组内
%   最佳与次佳的 Δ 和 GPR。模型输入与评估设置见附录 app:integration。
\caption{Performance comparison of existing motion-language evaluators on
STRIDE. We report matched-motion accuracy (acc), original-caption preference
under unrelated motions (bias), and normalized grounding gain $\Delta$;
acc and bias are reported as proportions. Macro GPR (\%) averages GPR over
the three structural axes, while Random Content serves as a content-matching
reference. Bold and underlined values denote the best and second-best
$\Delta$ and GPR within each dataset block, respectively.
Appendix~\ref{app:integration} details model inputs and evaluation settings.}
\label{tab:main}
\begin{center}
\scriptsize
\resizebox{\linewidth}{!}{% Mirror baselines for native TMR and SnapMoGen use the Fig. 9 matched scores.
% Unrelated-motion controls are unchanged; Delta and GPR were recomputed.
% Provenance: papers/motion_text_alignment/paper_working/table2_fig9_alignment/audit.json
\setlength{\tabcolsep}{1.6pt}
\renewcommand{\arraystretch}{0.92}
\begin{tabular}{lccccccccccccc}
\toprule
 & \multicolumn{3}{c}{Temporal Order} & \multicolumn{3}{c}{Mirror Reflection} & \multicolumn{3}{c}{Action Identity} & \multicolumn{3}{c}{Random Content} & \shortstack{Macro\\GPR (\%)$\uparrow$} \\
\cmidrule(lr){2-4}\cmidrule(lr){5-7}\cmidrule(lr){8-10}\cmidrule(lr){11-13}
Published evaluator & acc & bias & $\Delta\uparrow$ & acc & bias & $\Delta\uparrow$ & acc & bias & $\Delta\uparrow$ & acc & bias & $\Delta\uparrow$ &  \\
\midrule
\multicolumn{14}{l}{\textit{HumanML3D (native input)}} \\
HumanML3D evaluator~\citep{guo2022humanml3d} & 0.624 & 0.392 & 0.381 & 0.463 & 0.484 & -0.041 & 0.858 & 0.500 & 0.717 & 0.940 & 0.503 & 0.879 & 66.4 \\
TMR~\citep{petrovich2023tmr} & 0.704 & 0.494 & 0.416 & 0.728 & 0.527 & \underline{0.424} & 0.887 & 0.557 & 0.745 & 0.955 & 0.506 & 0.910 & 70.7 \\
CLaM~\citep{chen2024clam} & 0.726 & 0.441 & \textbf{0.510} & 0.737 & 0.519 & \textbf{0.454} & 0.925 & 0.538 & \textbf{0.837} & 0.978 & 0.490 & \textbf{0.956} & \textbf{73.9} \\
MoBERT~\citep{voas2023metric} & 0.726 & 0.495 & \underline{0.457} & 0.594 & 0.501 & 0.186 & 0.844 & 0.533 & 0.667 & 0.937 & 0.476 & 0.880 & 70.3 \\
ChronRet~\citep{fujiwara2024car} & 0.698 & 0.503 & 0.392 & 0.693 & 0.515 & 0.367 & 0.925 & 0.585 & \underline{0.818} & 0.970 & 0.514 & \underline{0.938} & \underline{72.4} \\
\addlinespace[1pt]
\multicolumn{14}{l}{\textit{SnapMoGen (long captions)}} \\
HumanML3D evaluator~\citep{guo2022humanml3d} & 0.541 & 0.503 & 0.076 & 0.513 & 0.504 & 0.018 & 0.564 & 0.505 & 0.119 & 0.748 & 0.516 & 0.478 & 52.5 \\
TMR~\citep{petrovich2023tmr} & 0.510 & 0.466 & 0.082 & 0.617 & 0.529 & 0.187 & 0.642 & 0.456 & \underline{0.342} & 0.789 & 0.520 & 0.560 & 57.7 \\
CLaM~\citep{chen2024clam} & 0.569 & 0.478 & \underline{0.174} & 0.600 & 0.488 & \underline{0.219} & 0.642 & 0.510 & 0.270 & 0.848 & 0.535 & \underline{0.672} & 56.9 \\
MoBERT~\citep{voas2023metric} & 0.552 & 0.501 & 0.102 & 0.534 & 0.499 & 0.070 & 0.547 & 0.493 & 0.106 & 0.733 & 0.535 & 0.425 & 55.8 \\
ChronRet~\citep{fujiwara2024car} & 0.568 & 0.496 & 0.143 & 0.599 & 0.543 & 0.123 & 0.593 & 0.549 & 0.098 & 0.816 & 0.502 & 0.631 & \underline{58.8} \\
SnapMoGen evaluator~\citep{guo2025snapmogen} & 0.965 & 0.518 & \textbf{0.927} & 0.728 & 0.519 & \textbf{0.435} & 0.838 & 0.529 & \textbf{0.656} & 0.998 & 0.500 & \textbf{0.995} & \textbf{74.7} \\
\bottomrule
\end{tabular}
}
\end{center}
\end{table}

\begin{figure}[htb]
\centering
\includegraphics[width=\linewidth]{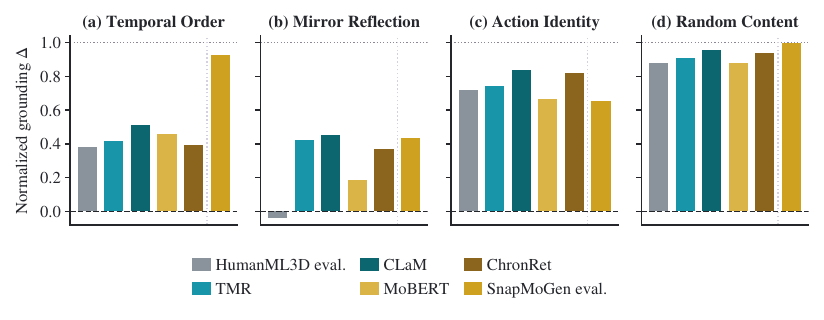}
% 【翻译】现有评测器在不同扰动轴上的性能比较。柱高为表 2 中的归一化接地分
%   Δ（越高越好），涵盖三条结构轴及随机内容参照。每个面板的前五个评测器在
%   HumanML3D 上测试，竖直分隔线右侧的 SnapMoGen 评测器在 SnapMoGen 上测试；
%   跨分隔线的柱高不构成同一测试集上的排名。横向虚线与点线分别表示 Δ=0
%   （相对于无关动作没有净增益）与 Δ=1（匹配动作下完全正确）。
\caption{Performance comparison of existing evaluators across perturbation axes.
Bars report normalized grounding $\Delta$ (higher is better) from
Table~\ref{tab:main} for the three structural axes and the Random Content
reference. In each panel, the first five evaluators are tested on
HumanML3D; the SnapMoGen evaluator, to the right of the vertical separator,
is tested on SnapMoGen. Comparisons across the separator are not
same-dataset rankings. The horizontal dashed and dotted lines mark
$\Delta=0$ (no gain over unrelated motion) and $\Delta=1$
(perfect matched-motion accuracy), respectively.}
\label{fig:acc-bias}
\end{figure}

% 【翻译】表 tab:main 显示，每个评测器在三条结构轴上的 Δ 均低于自身的随机内容
%   参照，揭示了内容匹配与结构区分之间的明显差距。图 fig:acc-bias 将各评测器在
%   原生数据集上的这一差距可视化。对多数评测器而言，镜像反射的 Δ 是三条结构轴中
%   最低的，表明左右区分是较普遍的薄弱环节。在 HumanML3D 上，时间顺序与镜像
%   反射尤为薄弱：即使表现最好的评测器，其 Δ 也分别仅为 0.510 和 0.454。
%   相比之下，动作身份更容易区分。在 SnapMoGen 上，常用的短描述训练评测器
%   面对该数据集的长描述表现更差。不过，SnapMoGen 评测器在时间顺序上具有明显
%   优势，Δ 达到 0.927。这可能得益于评测器训练中的动作重建约束：除跨模态对齐外，
%   该约束还要求从文本嵌入和动作嵌入分别重建对应的动作序列，可能促使表示保留
%   动作顺序，从而帮助检测时序不匹配。然而，其镜像反射与动作身份 Δ 仍远低于
%   自身的随机内容参照。
Table~\ref{tab:main} shows that every evaluator obtains lower $\Delta$
scores on all three structural axes than on its Random Content reference,
revealing a clear gap between content matching and structural
discrimination. Figure~\ref{fig:acc-bias} visualizes this gap for each
evaluator on its native dataset. For most evaluators, Mirror Reflection has
the lowest $\Delta$ of the three structural axes, indicating a recurring
weakness in left--right discrimination. On HumanML3D, Temporal Order and Mirror Reflection are
particularly weak: even the best-performing evaluator reaches $\Delta$
scores of only $0.510$ and $0.454$, respectively. Action Identity is
comparatively easier to distinguish. On SnapMoGen, commonly used
evaluators trained on short captions perform still worse with the
dataset's long captions. The SnapMoGen evaluator, however, shows a clear
advantage on Temporal Order, achieving a $\Delta$ of $0.927$. This may
benefit from the motion reconstruction constraint used during evaluator
training. In addition to cross-modal alignment, this constraint requires
both text and motion embeddings to reconstruct the corresponding motion
sequence, which may encourage the representations to preserve action
order and help detect temporal mismatches. Nevertheless, its Mirror
Reflection and Action Identity $\Delta$ scores remain well below its
Random Content reference.

% 【翻译】宏 GPR 进一步从样本层面补充上述 Δ 分析，衡量匹配动作是否普遍提高了原句相对于扰动句的得分优势。
% 所有受测评测器的宏 GPR 均低于 75%，得分最高的SnapMoGen 评测器也仅取得74.7%的分数。这表明，在相当一部分结构扰动句对上，现有评测器未能借助匹配动作增强原句相对于扰动句的得分优势。
Macro GPR complements the $\Delta$ analysis at the sample level by
measuring whether matched motions consistently increase the original
caption's score advantage over its perturbed counterpart. All tested
evaluators have a macro GPR below $75\%$, with even the highest-scoring
SnapMoGen evaluator reaching only $74.7\%$. This indicates that, for a
substantial fraction of structurally perturbed caption pairs, the tested
evaluators fail to use matched motions to increase the original caption's
score advantage over its perturbed counterpart.

% 【翻译】对生成模型和视觉语言模型的适配
%   测试还显示，高描述选择准确率可能伴随较强的无关动作描述偏好，而关键帧
%   视觉语言模型的结果也受到整体内容匹配能力不足的限制（接口设置与结果见附录 app:adapted-results）。
Evaluations on adapted generative and vision-language interfaces also show that high
caption-selection accuracy can coexist with strong caption preferences
under unrelated motions, while the keyframe-based VLM results are
limited by weak overall content matching, details are in Appendix~\ref{app:adapted-results}.

% ======================================================================
% 【翻译】结构错误为何会被漏掉？
\section{Why can structural errors be missed?}
\label{sec:anatomy}
% ======================================================================
% 【翻译】STRIDE 揭示了评测器在结构区分上的不足，但这些不足为何没有充分体现在
%   常用评测指标中？为了检验这一问题，我们进一步评估这些不足是否会影响评估器在动作生成任务中的评测结果。
%   我们将干预转向动作侧，首先检验常用评测指标能否识别动作出现结构错误。
%   随后，我们讨论对比训练的匹配目标为何可能未充分激励模型学习这些差异。
STRIDE exposes weaknesses in structural discrimination, but why are these
weaknesses not fully reflected in standard evaluation metrics? To examine
this question, we investigate whether these limitations affect the
evaluators' use in motion-generation evaluation. We shift the intervention
to the motion side and first test whether standard metrics detect
structural errors in the motions. We then discuss why the matching
objective in contrastive training may provide insufficient incentive to
learn these distinctions.

% 【翻译】基于检索的评测的局限性。
\subsection{Limitations of retrieval-based evaluations}
% 【翻译】为单独考察指标对结构错误的响应，我们从真实动作出发，保持描述不变，
%   对动作施加可控变换，再使用标准生成评测套件打分。这样可以避免不同生成
%   模型引入的其他质量差异干扰分析。表中比较原始真实动作、镜像动作、时间反转动作，以及打乱动作与
%   描述配对的对照。时间反转还会改变局部动力学，并不等同于 STRIDE 中只改变
%   事件先后顺序的文本扰动。
To isolate the metrics' response to structural errors, we start from
ground-truth motions, keep their captions fixed, and apply controlled
motion transformations before scoring the resulting pairs with the
standard generation-evaluation suite. This avoids confounding the
analysis with other quality differences between outputs from different
generators. Table~\ref{tab:audit} compares the original ground-truth
motions, mirrored motions, time-reversed motions, and a control with
shuffled motion-caption pairings. Time reversal also changes local
dynamics, so it is not equivalent to STRIDE's caption perturbations that
change only event order.

\begin{table}[ht]
% 【翻译】动作侧干预下标准动作生成评测指标的比较。我们以原始真实动作为参照，
%   比较镜像、时间反转与配对打乱三种条件。R@K 表示以小数比例报告的
%   R-Precision@K；FID 与 MM-Dist 分别为 Fréchet Inception Distance 和
%   动作—文本匹配距离。
\caption{Comparison of standard motion-generation metrics under motion-side
interventions. We compare mirroring, time reversal, and shuffled pairing
against the original ground-truth reference. R@$K$ denotes R-Precision
at $K$, reported as a proportion; FID and MM-Dist denote Fr\'echet Inception
Distance and motion-text matching distance, respectively.}
\label{tab:audit}
\begin{center}
\small
\begin{tabular}{lcccccc}
\toprule
Motion Condition & R@1$\uparrow$ & R@2$\uparrow$ & R@3$\uparrow$ & FID$\downarrow$ & MM-Dist$\downarrow$ & Diversity$\rightarrow$ \\
\midrule
Original Motion & 0.529 & 0.710 & 0.812 & 0.000 & 2.914 & 9.272 \\
Mirror Reflection & 0.523 & 0.713 & 0.815 & 0.020 & 2.904 & 9.777 \\
Time Reversal & 0.357 & 0.523 & 0.632 & 8.906 & 4.729 & 8.327 \\
Shuffled Pairing & 0.032 & 0.066 & 0.098 & 0.000 & 9.516 & 9.349 \\
\bottomrule
\end{tabular}

\end{center}
\end{table}

% 【翻译】表 tab:audit 显示，尽管匹配指标对时间反转和配对打乱有明显响应，
%   镜像反射后 R@1 仅从 0.529 变为 0.523，MM-Dist 甚至略优于
%   原始动作。这些结果表明，标准评测分数可能无法揭示动作与描述之间的左右
%   结构错配。在所评估的协议和数据集上，即使不能正确区分左右，也可能获得
%   较好的整体匹配分数，使这些错误未能体现在最终分数中。
Table~\ref{tab:audit} shows that, although the matching metrics respond
clearly to Time Reversal and Shuffled Pairing, mirroring changes R@1
only from $0.529$ to $0.523$, and even yields
a slightly better MM-Dist score than the original motions. These results
indicate that standard evaluation scores may fail to reveal left--right
structural mismatches between motions and captions. Under the evaluated
protocol and dataset, good overall matching scores can thus be obtained
without correctly distinguishing left from right, allowing these errors
to remain hidden in the final scores.

% 【翻译】基于对比学习的训练的局限性。
\subsection{Limitations of contrastive learning-based training}
% 【翻译】上述结果解释了结构缺陷为何可能在标准评测中被漏掉，却尚未解释这些
%   缺陷为何形成。一个可能的原因是，对于使用对比学习训练的动作—语言匹配器，训练
%   目标与负样本共同决定了模型必须区分哪些差异。目标要求匹配动作和描述优于
%   其他候选；如果候选主要在整体动作内容上不同，那么粗粒度内容匹配就可能
%   足以降低损失，而无需准确区分左右或事件顺序。
The preceding results explain how structural weaknesses can go
unnoticed in evaluation, but not why they arise. For contrastively
trained motion-language matchers, one possible explanation lies in the
combination of the objective and the available negatives. The objective
requires matched motions and captions to outrank alternatives. If those
alternatives differ mainly in overall action content, coarse content
matching may suffice to reduce the loss without resolving left--right
distinctions or event order.

% 【翻译】因此，关键不只是描述是否包含结构信息，而是训练时是否必须使用
%   这些信息才能区分正确配对与负样本。当内容相近、仅结构不同的候选不足时，
%   忽略结构也可能满足匹配目标。这是针对对比式匹配器的可能解释，而非本实验
%   已单独证实的训练成因，也不适用于所有受测模型的训练目标。
The issue is therefore not simply whether captions contain structural
information, but whether training requires that information to
distinguish a correct pair from its negatives. When alternatives with
similar content but different structure are missing, ignoring structure
may still satisfy the matching objective. This is a possible explanation
for contrastive matchers, not an isolated causal finding or a claim about
every tested model's training objective.

% ======================================================================
% 【翻译】一种简单的改进：结构困难负样本。
\section{A simple fix: structural hard negatives}
\label{sec:negtmr}
% ======================================================================
% 【翻译】上一节的分析提示，仅在整体内容上不同的负样本可能不足以要求模型区分
%   细微结构。我们因此加入内容相近、结构不同的困难负样本，检验其能否改善结构接地。这一改进是与评估器结构无关的，为了简化，我们将这一改进称为Neg Suite
The preceding analysis suggests that content-level negatives may not
require fine-grained structural distinctions. We therefore add
content-similar, structurally different hard negatives to test whether
they improve structural grounding. This approach requires no changes to
the evaluator architecture; we refer to it as Neg Suite.

\begin{figure}[!htbp]
\centering
\includegraphics[width=\linewidth]{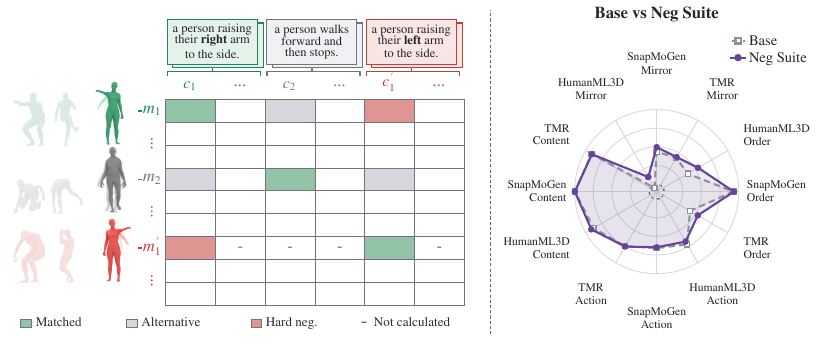}
% 【翻译】基于结构困难负样本微调。左：动作与描述组成的
%   正确配对、普通负样本和结构困难负样本。
%   动作侧新增监督只比较同一原始配对及其扰动版本。
%   右：比较 HumanML3D、TMR 和 SnapMoGen 的原始评测器（Base）与困难负样本训练
%   版本（Neg Suite）在三条结构轴及随机内容对照上的归一化接地分 Δ。
\caption{Finetuning with structural hard negatives.
Left: correct motion--caption pairs, ordinary negatives, and structural
hard negatives. The added motion-side supervision compares only an
original pair and its perturbed versions.
Right: normalized grounding $\Delta$ on the three structural axes and the
Random Content control for HumanML3D, TMR, and SnapMoGen, comparing the
original evaluators (Base) with their hard-negative-trained variants
(Neg Suite).}
\label{fig:targeted-negatives}
\end{figure}

% 【翻译】图 fig:targeted-negatives 展示了我们提出的结构困难负样本方法。我们保留原始对比学习中的普通负样本，分别加入扰动描述和扰动动作。对于新增的扰动描述和扰动动作，我们采用不对称设计。具体来说，扰动描述同时与原动作、Alternative动作以及扰动动作交互，而扰动动作仅与原描述和扰动描述交互。这一设计是为了让动作侧新增的梯度集中在“区分结构错误”，避免重复整体内容不匹配的训练信号。
Figure~\ref{fig:targeted-negatives} illustrates our structural hard-negative
approach. We retain the ordinary negatives used in standard contrastive
learning and introduce perturbed captions and motions. We use an
asymmetric design: perturbed captions are compared with the original
motion, other motions in the batch, and the corresponding perturbed
motion, whereas perturbed motions are compared only with their
corresponding original and perturbed captions. This design aims to focus
the additional motion-side gradients on distinguishing structural errors,
avoiding duplicating training signals for overall content mismatches.

\FloatBarrier

% 【翻译】我们在 HumanML3D、TMR 和 SnapMoGen 上检验该策略，微调版本依次记为
%   NegHML、NegTMR、NegSnap，比较归一化接地分 Δ 并检查检索是否退化。
%   负样本构造、变体损失和训练设置见附录 E。
We test HumanML3D, TMR, and SnapMoGen, comparing normalized grounding $\Delta$ and checking retrieval
degradation. Appendix~\ref{app:finetune-details} gives negative construction,
variant-specific losses, and training settings.

% 【翻译】图 fig:targeted-negatives 右侧展示逐轴 Δ。改进主要体现在时间顺序和镜像反射上。NegTMR 将这两条轴的 Δ 分别从
%   0.416 和 0.424 提高到 0.536 和 0.435；NegSnap 则分别从 0.927 和 0.435
%   提高到 0.935 和 0.496。TMR 与 SnapMoGen 的六个结构分数中有五个提高。
%   HumanML3D 评测器的镜像反射 Δ 也从 −0.041 提高到 0.127。这些结果表明，
%   加入结构困难负样本能够改善部分结构区分能力。
%   动作身份的收益则有限，部分评测器出现下降，可能因为这类困难负样本与普通
%   替代描述较为相似，提供的额外监督较少。
Figure~\ref{fig:targeted-negatives} shows per-axis $\Delta$.
The gains are most evident in Temporal Order and Mirror Reflection.
NegTMR raises $\Delta$ on these axes from $0.416$ and $0.424$ to $0.536$
and $0.435$, respectively; NegSnap raises them from $0.927$ and $0.435$
to $0.935$ and $0.496$. Across TMR and SnapMoGen, five of the six
structural scores improve. The HumanML3D evaluator's Mirror Reflection
$\Delta$ also rises from $-0.041$ to $0.127$. Structural hard negatives
can therefore improve some structural distinctions. Gains on Action
Identity are limited, with declines for some evaluators, possibly because
these hard negatives resemble ordinary alternative captions and provide
less additional supervision.

% ======================================================================
% 【翻译】相关工作。
\section{Related work}
\label{sec:related}
% ======================================================================
% 【翻译】动作—语言评测。HumanML3D 建立了文本到动作生成常用的评测
%   空间。MotionCLIP 把
%   动作对齐到 CLIP 文本空间；TMR 在 TEMOS 基础上强化跨模态检索，后续还有
%   跨数据集迁移研究；CLaM 提供 CLIP 式语言—动作评测器；SnapMoGen 使用 T5
%   文本表示构建长描述评测器；MoBERT 提供经人工判断校准的质量回归器；
%   MotionGPT 与 MotionLLM 通过条件生成反映动作与文本的对应关系。整段动作或分布层面的高分不保证结构区分能力，STRIDE 直接检验
%   评测器能否识别这些结构变化。
\textbf{Motion-language evaluation.} HumanML3D established
the common evaluation space for text-to-motion
generation~\citep{guo2022humanml3d}. MotionCLIP aligns motion with the CLIP text
space~\citep{tevet2022motionclip}; TMR extends TEMOS toward cross-modal
retrieval with a follow-up cross-dataset
study~\citep{petrovich2022temos,petrovich2023tmr,bensabath2024tmrpp}; CLaM
offers a CLIP-style evaluator~\citep{chen2024clam}, SnapMoGen a
long-caption T5 evaluator~\citep{guo2025snapmogen}, and MoBERT a
human-calibrated quality regressor~\citep{voas2023metric}.
MotionGPT and MotionLLM model motion-text correspondence through conditional
generation~\citep{jiang2023motiongpt,chen2024motionllm}.
High clip-level or distributional scores do not guarantee structural
discrimination; STRIDE directly tests whether evaluators recognize these
structural changes.

% 【翻译】细粒度与时间接地。CAR在原始描述与
%   事件乱序描述之间做检索，并把乱序描述作为训练负样本。
%   FineMoLA 从片段级监督学习帧—短语对齐；SGAR 通过身体部位概念对齐全身与
%   局部动作。在动作领域之外，视频—语言诊断集检验时间
%   敏感度，自监督识别工作则把时间方向与镜像手性当作可学习的信号。这些工作把
%   结构作为学习目标。STRIDE 在长短描述上对照三条结构轴与随机内容匹配，
%   分别控制数据侧文本偏置和无关动作下的描述偏好，并检验困难负样本能否在
%   不改变架构的情况下改善结构接地。
\textbf{Fine-grained and temporal grounding.} CAR retrieves the original description against an event-shuffled alternative and trains on such
negatives~\citep{fujiwara2024car}. FineMoLA learns
frame-phrase alignment from clip-level supervision~\citep{finemola2026},
and SGAR aligns full-body with part-level motion through body-part
concepts~\citep{zhang2025sgar}. Outside motion,
video-language diagnostics probe time
sensitivity~\citep{bagad2023testoftime,liu2024tempcompass,li2024vitatecs},
and self-supervised recognition treats temporal direction and mirror
chirality as learnable
signals~\citep{misra2016shuffle,wei2018arrow,lin2020chirality}. These works
make structure a learning target. STRIDE compares three structural axes
against Random Content matching across short and long captions, separately
controls data-side text-only bias and unrelated-motion caption preference,
and tests hard-negative training without architectural changes.

% 【翻译】组合性诊断与硬负样本。Winoground 用精心策划的图文对检验组合性；ARO
%   将关系、属性和顺序诊断扩展到大规模测试，并表明定向硬负样本可以改善对比式
%   模型。后继探针覆盖组合式检索、空间关系与对象—属性清单。SugarCrepe 随后
%   证明扰动基准可被纯文本先验攻破，SugarCrepe++ 加入保义换词下的敏感度测试，
%   呼应 VQA 的语言先验配平。基于这些组合性与偏置诊断，STRIDE 检验评测器的
%   判断是否依赖动作结构，以及常用生成指标能否识别结构错配。
\textbf{Compositionality diagnostics and hard-negative learning.} Winoground
tests composition with curated pairs, and ARO scales relation, attribution,
and order diagnostics while showing that targeted negatives improve
contrastive models~\citep{thrush2022winoground,yuksekgonul2023aro}; later
probes cover compositional retrieval, spatial relations, and
object-attribute
checklists~\citep{ma2023crepe,ray2023cola,kamath2023whatsup,zhao2022vlchecklist}.
SugarCrepe then demonstrates that text-only priors can compromise
perturbation benchmarks, and SugarCrepe++ adds sensitivity tests under
meaning-preserving lexical
variation~\citep{hsieh2023sugarcrepe,dumpala2024sugarcrepepp}, echoing
language-prior balancing in VQA~\citep{goyal2017vqa2}. Building on these
compositionality and bias diagnostics, STRIDE asks whether evaluator
judgments reflect motion structure,
and whether standard generation metrics detect structural mismatches.

% 【翻译】结论。
\section{Conclusion}
\label{sec:conclusion}
% 【翻译】在本研究中，我们提出STRIDE，揭示了动作—语言评测器在内容匹配与结构区分之间的能力差距。
%   在长短描述上，所有受测的已发布评测器在结构轴上的接地分均低于各自的
%   随机内容参照. 我们实验结果表明标准生成评测指标可能漏掉左右结构错误。无关动作下仍存在的描述偏好也说明，仅凭
%   选对描述不足以判断模型是否利用了动作。我们认为，这种结构能力缺口的一个可能原因是：对于对比式匹配器，
%   当负样本主要在整体内容上不同时，训练目标可能并不要求区分细微结构差异。
%   基于这一假设，我们加入文本侧和动作侧的结构困难负样本检验针对性监督的作用，观察到
%   时间顺序和镜像反射能力得到改善。未来可进一步研究训练目标与结构接地的关系，
%   并将测试扩展至更多模型、动作时长和身体部位等结构。动作—语言评测应在检索
%   和分布指标之外加入针对性的结构测试. 它们不能仅从整体分数推断结构理解能力，这能够进一步揭示MoLMs的优势和缺陷，也是理解它们适用场景的关键。
In this work, we introduce STRIDE, revealing a gap between content matching
and structural discrimination in motion-language evaluators. Across short
and long captions, all tested released evaluators obtain lower grounding
scores on structural axes than on their Random Content reference. Our
experiments show that standard generation metrics can miss left--right
structural errors. Persistent caption preferences under unrelated motions
also show that selecting the correct caption alone does not establish
that a model uses motion. One possible explanation for this structural
capability gap is that, for contrastive matchers, the training objective
may not require fine-grained structural distinctions when negatives
differ mainly in overall content. Motivated by this hypothesis, we add
structural hard negatives on the caption and motion sides to test targeted
supervision, observing improvements in Temporal Order and Mirror
Reflection. Future work could further examine the relationship between
training objectives and structural grounding, and extend the tests to
more models and structures such as motion duration and body-part
structure. Motion-language evaluation should complement retrieval and
distributional metrics with targeted structural tests. Structural understanding should not be inferred from aggregate scores alone. Such tests can
further reveal the strengths and weaknesses of motion-language models
and are key to understanding where these models are applicable.

% ======================================================================
% \clearpage
% 【翻译】AI 使用声明。
\subsubsection*{AI usage statement}
% 【翻译】本研究使用 AI 工具辅助写作与润色、文献检索，以及合成描述数据的生成。
%   具体而言，AI 辅助了中英文翻译、语言润色、相关工作发现与参考文献核查。
%   在数据构建中，生成式 AI 仅用于描述文本：大语言模型生成时间顺序扰动，
%   人工审计评估构建质量与纯文本偏好。作者对研究计划的制定和执行，以及最终
%   研究决策、解释、参考文献和稿件内容承担全部责任。
We used AI tools to assist with writing and polishing, literature retrieval,
and generating synthetic caption data. Specifically, AI assisted with
Chinese--English translation, language polishing, related-work discovery,
and reference checking. In dataset construction, generative AI was used
only for caption text: a large language model generated Temporal Order
perturbations, and human audits assessed construction quality and text-only
preference. The authors take full responsibility for developing the research
plan, carrying it out, and the final research decisions, interpretations,
references, and manuscript content.

% 【翻译】伦理声明。
\subsubsection*{Ethics statement}
% 【翻译】我们通过 AMASS 获取 HumanML3D，并获取 SnapMoGen，分别遵循其学术许可。
%   两个数据集均包含骨架化动作，不含可识别个人身份的信息。
%   我们发布的产物只包含描述文本、扰动文本、片段 id 与代码——不再分发任何
%   动作数据；复现需要用户自行在原始发布方处同意相应许可。人工标注仅涉及
%   非敏感的动作—描述判断。五名 22-30 岁的自愿标注者在参与前已了解研究目的
%   及其回答的用途。研究未收集可识别身份或敏感信息。根据机构规范咨询后，
%   该标注流程被认定为无需 IRB 审查。
We obtained HumanML3D via AMASS and SnapMoGen under their respective academic
licenses. Both datasets contain skeletonized motions without personally
identifiable information. Released artifacts contain only captions, perturbed captions,
clip identifiers, and code. No motion data is redistributed; reproduction
requires agreeing to the original licenses with the respective providers.
Human annotation was limited to non-sensitive judgments of motion-caption
pairs. Five volunteer annotators aged 22--30 were informed of the study purpose
and the intended use of their responses before participation. No identifying or
sensitive information was collected. Following consultation under our
institutional guidelines, the protocol was determined not to require IRB review.

% 【翻译】可重复性声明。
\subsubsection*{Reproducibility statement}
% 【翻译】论文接收后，我们将公开 STRIDE 基准数据，以及用于复现本文实验和
%   构建基准的代码。发布内容包括全部 5,869 个句对的文本与动作片段标识符、
%   逐对的双向镜像分数网格、训练负样本，以及构建、评测、训练和分析脚本。
%   原始动作数据须按原数据集许可向其发布方获取，不由我们重新分发。
Upon acceptance, we will publicly release the STRIDE benchmark data and
the code for reproducing the experiments and constructing the benchmark.
The release will include all $5{,}869$ caption pairs with motion clip
identifiers, per-pair bidirectional mirror score grids, training negatives,
and construction, evaluation, training, and analysis scripts.
The original motion data must be obtained from the dataset providers under
their respective licenses and will not be redistributed.

\bibliography{iclr2027_conference}
\bibliographystyle{iclr2027_conference}

\newpage
\appendix
\FloatBarrier
% 【翻译】基准构建与覆盖
\section{Benchmark construction and coverage}
\label{app:datasheet}

\subsection{Construction steps}
\label{app:record}

\begin{figure}[ht]
\centering
\includegraphics[width=\linewidth]{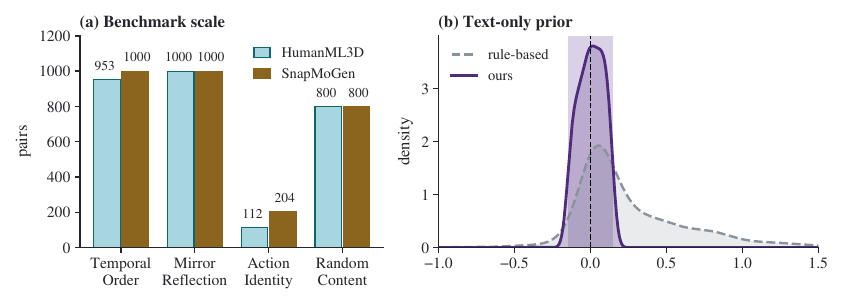}
% 【翻译】基准组成与似然配平。(a) 各轴、各语料的句对数量。(b) 控制前后的
%   Δℓℓ 平滑密度。正值表示偏好原句，阴影标出接受区间。
\caption{Benchmark composition and likelihood balancing. (a) Pair counts
by axis and corpus. (b) Smoothed $\Delta\ell\ell$ densities before and after
control. Positive values indicate preference for the original
caption; the shaded region marks the acceptance interval.}
\label{fig:dataset}
\end{figure}

% 【翻译】图 fig:dataset 按语料与扰动轴汇总 5,869 个句对。结构句对按以下步骤构造。
Figure~\ref{fig:dataset} summarizes the $5{,}869$ pairs by corpus and axis.
We construct structural pairs as follows.

% 【翻译】生成编辑。时间顺序使用大语言模型改写来反转事件顺序，本文中使用开源的Qwen3-VL-8B模型，要求保留全部动作、尽量少改措辞且文本流畅。镜像反射可以根据词直接进行替换，如交换 left/right、clockwise/counterclockwise等。
%   动作身份从视觉上可区分的动作词对中替换一个动词，避免run替换成jog这种难以区分的例子。
\paragraph{Generate edits.} Temporal Order uses an LLM to rewrite captions
with the event order reversed. We use the open-source
Qwen3-VL-8B~\citep{bai2025qwen3vl}, instructing it to retain all actions,
minimize wording changes, and produce fluent text. Mirror Reflection can
be constructed through direct word substitutions, such as swapping
left/right or clockwise/counterclockwise. Action Identity replaces one
verb using visually distinguishable action pairs, avoiding substitutions
such as replacing run with jog, which are difficult to distinguish visually.
% 【翻译】检查候选。时间顺序编辑须保留实词并通过结构检查。动作身份排除源动作
%   或目标动作重复出现的候选；人工检查流畅性、动作是否不同且互斥，以及是否只改变一个谓词。

\paragraph{Check candidates.} Order edits must preserve content words and pass
structural checks. Action edits reject candidates in which the source or
target action appears more than once. Human reviewers check fluency,
whether the actions are distinct and mutually exclusive, and whether
only one predicate has changed.
% 【翻译】筛选与配平。为了降低text-only先验的影响，所有结构句对须满足|Δℓℓ|≤0.15 nats/token。
%   然后动作身份需要额外进行配平，保证每组动作词对的两个替换方向数量相等，避免某个动作词总是出现在正确描述或错误描述中，形成词汇捷径。

\paragraph{Filter and balance.} To reduce the influence of text-only
priors, we retain only structural pairs satisfying
$|\Delta\ell\ell|\leq0.15$ nats/token. We then balance Action Identity
pairs so that each action pair has equal numbers of examples in both
replacement directions. This prevents an action word from consistently
appearing in either the correct or the incorrect caption and providing
a lexical shortcut.

% 【翻译】词汇与动作覆盖
\subsection{Vocabulary and motion coverage}
\label{app:coverage}

% 【翻译】图 fig:coverage 汇总了词汇与动作覆盖情况。每个语料使用 5,000 段动作，以 14 项按身高归一化的运动学统计量描述。我们使用t-SNE可视化两套语料的数据分布，两个数据集
%   在投影中大幅重叠，其中SnapMoGen 分布更均匀。图 fig:coverage(c) 中两套语料的平滑覆盖区域大幅重叠，仅在边缘存在
%   局部差异，表明它们在该投影中覆盖了相近的运动学区域。
%   图 fig:coverage(d) 中Occ. 统计非空格数，Eff. 以网格样本分布的熵取指数衡量有效覆盖，Tail 统计
%   落入两套语料合计仅含一至两条样本的稀疏格子中的样本比例。
%   相同采样量下，SnapMoGen 的占用格数略多（920 对 870），有效格数也更高
%   （708 对 599），说明其在这些区域内的分布更均匀。
Figure~\ref{fig:coverage} summarizes vocabulary and motion coverage.
We represent $5{,}000$ clips per corpus with $14$ body-height-normalized
kinematic statistics and visualize both corpora in a shared t-SNE
projection. Their distributions overlap broadly, with SnapMoGen more
evenly spread. The smoothed coverage regions in
Figure~\ref{fig:coverage}(c) overlap substantially, with local differences
at their boundaries, indicating that the corpora cover similar kinematic
regions in this projection. In Figure~\ref{fig:coverage}(d), Occ. counts
nonempty cells, Eff. measures effective coverage through exponentiated
grid-distribution entropy, and Tail measures the sample share in sparse cells
containing only one or two samples across both corpora. With equal sample
sizes, SnapMoGen occupies slightly more cells ($920$ vs.\ $870$) and has
more effective cells ($708$ vs.\ $599$), indicating a more even spread
within these regions.

\FloatBarrier
% 【翻译】人工审计
\section{Human audits}
\label{app:humanaudit}

% 【翻译】我们执行两类人工审计。Text-only preference audit检验：不看动作时，标注者是否仍倾向于选择原句。五位标注者被要求在不看动作时选择纯文本句对，可选择原句、扰动句、
%   “同样可能”或“无法判断”。结果显示最终组成的bench原句偏好率为 56.3%。
We perform two types of human audits. The text-only preference audit examines whether annotators still prefer the original caption
without seeing the motion. Five annotators assess text-only caption pairs,
choosing the original caption, the perturbed caption, \emph{equally likely},
or \emph{uncertain}. The audit yields an original-caption preference rate
of $56.3\%$ for the resulting benchmark.

% 【翻译】构造质量审计检查描述编辑是否满足预期的构造要求。所有判断须一致确认描述流畅、只改变
%   一项结构，并在呈现动作时确认变化可由视觉区分。
The construction quality audit checks whether caption edits meet the intended construction
requirements. All judgments must agree that the captions are fluent,
only one structural element has changed, and, when motion is shown,
the change is visually distinguishable.

\FloatBarrier
% 【翻译】评估实现
\section{Evaluation implementation}
\label{app:integration}

% 【翻译】评测器使用各自发布的检查点，采用原生预处理。
Evaluators use their released checkpoints and native preprocessing.

\begin{enumerate}
% 【翻译】HumanML3D 评测器：使用在 HumanML3D 上训练的 GloVe--双向 GRU 版本。实现来自作者的 text-to-motion 仓库。
\item HumanML3D evaluator~\citep{guo2022humanml3d}: We use the
HumanML3D-trained GloVe--bidirectional-GRU evaluator.
The implementation originates from
\url{https://github.com/EricGuo5513/text-to-motion}.

% 【翻译】TMR：使用在 HumanML3D 上训练、采用 Guo 等人动作特征的版本。代码及预训练权重下载入口见官方仓库。
\item TMR~\citep{petrovich2023tmr}: We use the HumanML3D-trained variants. Code and weights are available through
\url{https://github.com/Mathux/TMR}.

% 【翻译】CLaM：使用发布的 HumanML3D-synthesis 评测器。
%   代码与权重下载入口见官方仓库。
\item CLaM~\citep{chen2024clam}: We use the released HumanML3D-synthesis
evaluator.
Code and checkpoint downloads are provided at
\url{https://github.com/SheldongChen/CLaM}.

% 【翻译】MoBERT：使用发布的 std_bpe2000 主评测器。代码与权重见官方仓库。
\item MoBERT~\citep{voas2023metric}: We use the released
\texttt{std\_bpe2000} primary evaluator. Code and weights are available through
\url{https://github.com/jvoas655/MoBERT}.

% 【翻译】ChronRet（CAR）：使用在 HumanML3D 上训练的版本。代码见官方仓库，发布权重见 Hugging Face。
\item ChronRet (CAR)~\citep{fujiwara2024car}: We use the HumanML3D-trained
variants.
Code is available at \url{https://github.com/line/ChronAccRet}, with
released weights at \url{https://huggingface.co/line-corporation/ChronAccRet}.

% 【翻译】SnapMoGen 评测器：使用在 SnapMoGen 上训练的六层评测器
%   eval_klde-5_late-5_nlayer6_norm，文本骨干为 T5-v1.1-base。代码与检查点下载入口见官方仓库。
\item SnapMoGen evaluator~\citep{guo2025snapmogen}: We use the
SnapMoGen-trained six-layer evaluator
\texttt{eval\_klde-5\_late-5\_nlayer6\_norm}, with a T5-v1.1-base
text backbone.
Code and checkpoint downloads are provided at
\url{https://github.com/snap-research/SnapMoGen}.
\end{enumerate}

% 【翻译】对五个基于 HumanML3D 训练的评测器，我们用发布的运动学代码解码
%   SnapMoGen 的 148 维、30 fps、24 关节输入，将其映射为 22 个关节，重采样至
%   20 fps，并编码为 263 维 HumanML3D 特征。
For the five HumanML3D-trained evaluators, SnapMoGen's $148$-dimensional,
$30$-fps, $24$-joint input is decoded with released kinematics, mapped to
$22$ joints, resampled to $20$ fps, and encoded as $263$-dimensional
HumanML3D features.

\FloatBarrier
% 【翻译】额外实验
\section{Additional experiments}
\label{app:evaluation-results}

% 【翻译】已发布评测器的接地结果
\subsection{Released-evaluator grounding results}
\label{app:evaluator-results}

% 【翻译】表 tab:gpr-full 以逐轴 GPR 补充正文的接地结果。
Table~\ref{tab:gpr-full} complements the main-text grounding results with
per-axis GPR.
\begin{table}[ht]
% 【翻译】逐轴 GPR（百分比；随机水平为 50%）。加粗与下划线分别标记各语料内
%   最佳与次佳值。† 表示固定适配器的跨域迁移。
\caption{Per-axis GPR (\%; chance $50\%$). Bold and underline mark the best
and second-best values within each corpus. $\dagger$ denotes fixed-adapter
cross-domain transfer.}
\label{tab:gpr-full}
\begin{center}
\scriptsize
\setlength{\tabcolsep}{5.0pt}
\renewcommand{\arraystretch}{0.92}
\begin{tabular}{lcccc}
\toprule
Published evaluator & \shortstack{Temporal\\Order $\uparrow$} & \shortstack{Mirror\\Reflection $\uparrow$} & \shortstack{Action\\Identity $\uparrow$} & Macro$\uparrow$ \\
\midrule
\multicolumn{5}{l}{\textit{HumanML3D (native input)}} \\
HumanML3D evaluator~\citep{guo2022humanml3d} & 66.9 & 48.4 & \underline{84.0} & 66.4 \\
TMR~\citep{petrovich2023tmr} & 65.5 & \underline{66.6} & 80.2 & 70.7 \\
CLaM~\citep{chen2024clam} & \underline{68.1} & 64.0 & \textbf{89.6} & \textbf{73.9} \\
MoBERT~\citep{voas2023metric} & \textbf{69.7} & 58.5 & 82.5 & 70.3 \\
ChronRet~\citep{fujiwara2024car} & 67.4 & \textbf{66.9} & 83.0 & \underline{72.4} \\
\addlinespace[1pt]
\multicolumn{5}{l}{\textit{SnapMoGen (long captions)}} \\
HumanML3D evaluator\textsuperscript{$\dagger$}~\citep{guo2022humanml3d} & 53.6 & 49.4 & 54.4 & 52.5 \\
TMR\textsuperscript{$\dagger$}~\citep{petrovich2023tmr} & 54.5 & 57.8 & \underline{60.8} & 57.7 \\
CLaM\textsuperscript{$\dagger$}~\citep{chen2024clam} & 58.9 & 57.7 & 53.9 & 56.9 \\
MoBERT\textsuperscript{$\dagger$}~\citep{voas2023metric} & 55.5 & 52.3 & 59.6 & 55.8 \\
ChronRet\textsuperscript{$\dagger$}~\citep{fujiwara2024car} & \underline{59.0} & \underline{58.0} & 59.3 & \underline{58.8} \\
SnapMoGen evaluator~\citep{guo2025snapmogen} & \textbf{87.1} & \textbf{67.5} & \textbf{69.6} & \textbf{74.7} \\
\bottomrule
\end{tabular}

\end{center}
\end{table}

\FloatBarrier
% 【翻译】适配的生成模型与视觉语言模型
\subsection{Adapted generative and vision-language models}
\label{app:adapted-results}

\label{app:m2t-adapters}
\label{app:vlm-interface}

% 【翻译】本节汇总适配模型的接口设置与实验结果。
This section presents the interface settings and results for adapted models.

% 【翻译】动作到文本似然适配
\paragraph{Motion-to-text likelihood adapters.}

% 【翻译】我们通过动作到文本（M2T）适配接口测试 MotionGPT、MotionGPT3、
%   MG-MotionLLM 和 IRG-MotionLLM。给定动作，我们逐 token 计算模型对已有描述的
%   预测概率，并以平均对数概率作为描述得分。每一步均使用描述中实际出现的前文，
%   而非模型自行生成的前文；计入结束 token，但不计提示词和填充 token。
MotionGPT, MotionGPT3, MG-MotionLLM, and IRG-MotionLLM are tested through
motion-to-text (M2T) adapters~\citep{jiang2023motiongpt,zhu2025motiongpt3humanmotionsecond,wu2025mg,li2025irg-motionllm}.
Given a motion, we compute the probability of each token in an existing
caption and use the mean log-probability as its score. At each step, we
condition on the caption's actual preceding tokens rather than tokens
generated by the model. We include the end token but exclude prompt and
padding tokens.

% 【翻译】视觉语言强制选择接口
\paragraph{Vision-language forced-choice interface.}

% 【翻译】Qwen-VL 返回类别选择，用匹配与无关动作下的选择率计算 Δ。
Qwen-VL returns a categorical choice; its $\Delta$ uses the matched and
unrelated-motion choice rates.

% 【翻译】VLM 接收从左到右排列、附带根轨迹的六帧均匀采样骨架。提示词说明
%   时间顺序并要求选择 A/B，描述位置随机排列。对照条件将动作替换为同一语料、
%   同一轴内动作 ID 不同的片段。
VLMs receive six uniformly spaced skeleton frames with root trajectories,
ordered left to right. Prompts specify time order and request A/B, with
randomized caption positions. Controls substitute a different motion ID
within the same corpus--axis cell. 
% 【翻译】结果与分析。表 tab:adapted-interfaces 汇总上述两类适配接口的结果。
\paragraph{Results and analysis.} Table~\ref{tab:adapted-interfaces}
summarizes results for the two interfaces above.

% 【翻译】MG-MotionLLM 在似然适配器中取得最高结构得分，IRG 第一阶段的分数
%   接近或低于零。Qwen-VL 较弱的随机内容匹配限制了对其结构得分的解读。
MG-MotionLLM leads the likelihood adapters on structural scores, while
IRG Stage-1 is near or below zero. Weak Random Content matching limits
interpretation of Qwen-VL's structural scores.

\begin{table}[ht]
% 【翻译】适配的非评测器接口。(a) M2T 模型使用长度归一化的 log p(c|m)；
%   (b) Qwen-VL 使用六关键帧强制选择。指标定义同表 tab:main；宏 GPR 为
%   三条结构轴的均值（零假设值为 50%）。面板 (b) 的随机内容未设置无关动作
%   对照，因此只报告 acc。(a) 中加粗标记最佳 acc、Δ 和 GPR，下划线标记
%   次佳 Δ 和 GPR；(b) 中加粗标记各语料内最佳 acc 与 Δ，按四舍五入前的数值排名。
\caption{Adapted non-evaluator interfaces. (a) M2T models use length-normalized
$\log p(c\mid m)$; (b) Qwen-VL uses six-keyframe forced choice. Metrics follow
Table~\ref{tab:main}; Macro GPR averages three structural axes (null $50\%$).
Panel (b) reports only acc for Random Content. In (a), bold marks best acc, $\Delta$, and GPR, and underline marks
second-best $\Delta$ and GPR. In (b), bold marks best acc and $\Delta$ per
corpus, ranked before rounding.}
\label{tab:adapted-interfaces}
\begin{center}
\textbf{(a) M2T Likelihood Adapters on HumanML3D}\par\vspace{2pt}
\resizebox{\columnwidth}{!}{\setlength{\tabcolsep}{1.6pt}
\renewcommand{\arraystretch}{0.96}
\begin{tabular}{lccccccccccccc}
\toprule
 & \multicolumn{3}{c}{Temporal Order} & \multicolumn{3}{c}{Mirror Reflection} & \multicolumn{3}{c}{Action Identity} & \multicolumn{3}{c}{Random Content} & \shortstack{Macro\\GPR (\%)$\uparrow$} \\
\cmidrule(lr){2-4}\cmidrule(lr){5-7}\cmidrule(lr){8-10}\cmidrule(lr){11-13}
M2T likelihood adapter & acc & bias & $\Delta\uparrow$ & acc & bias & $\Delta\uparrow$ & acc & bias & $\Delta\uparrow$ & acc & bias & $\Delta\uparrow$ &  \\
\midrule
MotionGPT~\citep{jiang2023motiongpt} & 0.902 & 0.894 & 0.081 & 0.730 & 0.514 & \underline{0.445} & 0.764 & 0.679 & 0.265 & 0.621 & 0.488 & 0.261 & 62.9 \\
MotionGPT3~\citep{zhu2025motiongpt3humanmotionsecond} & 0.935 & 0.873 & \underline{0.492} & 0.704 & 0.526 & 0.374 & 0.783 & 0.557 & \underline{0.511} & \textbf{0.767} & 0.518 & \textbf{0.516} & \underline{68.9} \\
MG-MotionLLM~\citep{wu2025mg} & \textbf{0.982} & 0.960 & \textbf{0.541} & \textbf{0.762} & 0.514 & \textbf{0.510} & \textbf{0.868} & 0.557 & \textbf{0.702} & 0.747 & 0.514 & \underline{0.480} & \textbf{69.7} \\
IRG-MotionLLM (Stage-1)~\citep{li2025irg-motionllm} & 0.953 & 0.955 & -0.048 & 0.475 & 0.475 & 0.000 & 0.726 & 0.726 & 0.000 & 0.541 & 0.522 & 0.041 & 50.5 \\
\bottomrule
\end{tabular}
}
\par\vspace{5pt}
\textbf{(b) Qwen-VL Keyframe Forced-Choice Adapters}\par\vspace{2pt}
\scriptsize
\setlength{\tabcolsep}{2.6pt}
\renewcommand{\arraystretch}{0.96}
\begin{tabular}{llcccccccccc}
\toprule
 &  & \multicolumn{3}{c}{Temporal Order} & \multicolumn{3}{c}{Mirror Reflection} & \multicolumn{3}{c}{Action Identity} & \shortstack{Random Content\\accuracy $\uparrow$} \\
\cmidrule(lr){3-5}\cmidrule(lr){6-8}\cmidrule(lr){9-11}
Caption regime & Model & acc & bias & $\Delta\uparrow$ & acc & bias & $\Delta\uparrow$ & acc & bias & $\Delta\uparrow$ &  \\
\midrule
\multirow{2}{*}{HumanML3D} & Qwen3-VL-8B & 0.620 & 0.567 & 0.123 & 0.513 & 0.480 & \textbf{0.064} & 0.616 & 0.554 & 0.140 & \textbf{0.687} \\
  & Qwen2-VL-7B & \textbf{0.753} & 0.660 & \textbf{0.275} & \textbf{0.547} & 0.553 & -0.015 & \textbf{0.670} & 0.554 & \textbf{0.260} & 0.680 \\
\addlinespace[1pt]
\multirow{2}{*}{SnapMoGen} & Qwen3-VL-8B & \textbf{0.807} & 0.747 & \textbf{0.237} & \textbf{0.540} & 0.493 & \textbf{0.092} & \textbf{0.613} & 0.540 & \textbf{0.159} & \textbf{0.633} \\
  & Qwen2-VL-7B & 0.767 & 0.740 & 0.103 & 0.500 & 0.513 & -0.027 & 0.607 & 0.573 & 0.078 & 0.520 \\
\bottomrule
\end{tabular}

\end{center}
\end{table}

\FloatBarrier
% 【翻译】结构困难负样本训练与结果
\section{Structural hard-negative training and results}
\label{app:finetune-details}

% 【翻译】Neg Suite 在普通对比学习负样本之外，引入描述侧与动作侧的结构困难
%   负样本，以加强对细粒度结构差异的监督。我们保留普通负样本，并使用内容相近、
%   结构不同的描述或动作提供额外监督。
Neg Suite introduces structural hard negatives on the caption and motion
sides, in addition to ordinary contrastive negatives, to strengthen
supervision for fine-grained structural distinctions. We retain ordinary
negatives and use captions or motions with similar content but different
structure to provide additional supervision.

% 【翻译】损失与训练设置
\subsection{Losses and training settings}
\label{app:training}
\label{app:neghml-training}

% 【翻译】描述侧负样本覆盖时间顺序、镜像反射和动作身份。针对动作身份的
%   附加描述配对损失及其与混合对比损失的组合如下：
Caption-side negatives cover Temporal Order, Mirror Reflection, and
Action Identity. The additional caption-pair loss for Action Identity
and its combination with the pooled contrastive loss are:
\[
\mathcal{L}_{\mathrm{pair}}
=\frac{1}{|\mathcal{A}|}\sum_{i\in\mathcal{A}}
\operatorname{softplus}\!\left(
\frac{s(m_i,c'_i)-s(m_i,c_i)}{\tau}\right),\qquad
\mathcal{L}=\mathcal{L}_{\mathrm{pool}}+0.5\mathcal{L}_{\mathrm{pair}}.
\]
% 【翻译】A 为动作词对集合，c_i 和 c'_i 为正确及动作编辑描述，τ=0.1；
%   L_pool 为混合负样本的对比损失。
Here $\mathcal{A}$ indexes action pairs, $c_i$ and $c'_i$ are correct and
action-edited captions, $\tau=0.1$, and $\mathcal{L}_{\mathrm{pool}}$ is
the pooled contrastive loss.

% 【翻译】所有 Neg Suite 变体共用下列优化设置。
All Neg Suite variants share the optimization settings below.
\begin{center}
\small
\begin{tabular}{@{}ll@{}}
\toprule
Setting & Value \\
\midrule
Optimizer & AdamW \\
Learning Rate / Weight Decay & $10^{-5}$ / $10^{-4}$ \\
Batch Size / Epochs & 32 / 2 \\
Temperature / Gradient-Norm Limit & 0.1 / 1.0 \\
Random Seed & 42 \\
\bottomrule
\end{tabular}
\end{center}

% 【翻译】干预实验详细结果
\subsection{Detailed intervention results}
\label{app:intervention-results}

% 【翻译】混合训练提高了镜像审计中的平均对角交互：TMR 从 0.094 提高到 0.297，
%   SnapMoGen 从 0.130 提高到 0.230。
Pooled training increases mean diagonal interaction in the mirror audit:
TMR from $0.094$ to $0.297$, SnapMoGen from $0.130$ to $0.230$.

% 【翻译】NegHML 的镜像描述选择准确率为 0.600、Δ=0.127，原评测器分别为
%   0.463 和 −0.041。在相同的 948 个双向镜像对上，对角准确率从 48.05%
%   提高到 71.04%。原句与镜像句条件下的动作侧准确率分别为 61.92% 和 61.71%。
NegHML selects the correct mirror caption at $0.600$ accuracy and
$\Delta=0.127$, compared with $0.463$ and $-0.041$ for the original
evaluator. On the same $948$ bidirectional pairs, diagonal accuracy
rises from $48.05\%$ to $71.04\%$. Motion-side accuracy is $61.92\%$
under the original captions and $61.71\%$ under the mirrored captions.

\begin{table}[htb]
% 【翻译】结构困难负样本干预。指标定义同表 tab:main。加粗标记每组基线—微调模型中较优的 acc
%   与 Δ，bias 不排名。MedR 为检索中位排名，越低越好。
%   所有变体均使用描述侧负样本，仅 NegHML 额外使用成对动作监督；‡ 标记使用
%   独立动作身份检查点（NegTMR-A、NegSnap-A）的结果，其余 NegTMR/NegSnap 结果来自混合训练检查点。
\caption{Structural-hard-negative interventions. Metrics follow
Table~\ref{tab:main}. Bold marks better acc and $\Delta$ within each
base--finetuned pair; bias is not ranked. MedR is median retrieval rank
(lower is better). All variants use caption negatives, only NegHML adds
paired motion supervision, and $\ddagger$ marks separate Action Identity
checkpoints (NegTMR-A and NegSnap-A), with other NegTMR/NegSnap results
from pooled checkpoints.}
\label{tab:negtmr}
\begin{center}
\scriptsize
\resizebox{\linewidth}{!}{\setlength{\tabcolsep}{2.2pt}
\renewcommand{\arraystretch}{0.96}
\begin{tabular}{lccccccccccccc}
\toprule
 & \multicolumn{3}{c}{Temporal Order} & \multicolumn{3}{c}{Mirror Reflection} & \multicolumn{3}{c}{Action Identity} & \multicolumn{3}{c}{Random Content} & \multicolumn{1}{c}{Retrieval} \\
\cmidrule(lr){2-4}\cmidrule(lr){5-7}\cmidrule(lr){8-10}\cmidrule(lr){11-13}\cmidrule(lr){14-14}
Model & acc$\uparrow$ & bias & $\Delta\uparrow$ & acc$\uparrow$ & bias & $\Delta\uparrow$ & acc$\uparrow$ & bias & $\Delta\uparrow$ & acc$\uparrow$ & bias & $\Delta\uparrow$ & MedR$\downarrow$ \\
\midrule
HumanML3D evaluator & 0.624 & 0.392 & 0.381 & 0.463 & 0.484 & -0.041 & \textbf{0.858} & 0.500 & \textbf{0.717} & 0.940 & 0.503 & 0.879 & 77 \\
~~+NegHML (ours) & \textbf{0.732} & 0.418 & \textbf{0.540} & \textbf{0.600} & 0.542 & \textbf{0.127} & 0.849 & 0.538 & 0.673 & \textbf{0.958} & 0.524 & \textbf{0.912} & 66 \\
\midrule
TMR & 0.704 & 0.494 & 0.416 & 0.728 & 0.527 & 0.424 & \textbf{0.887} & 0.557 & 0.745 & \textbf{0.955} & 0.506 & \textbf{0.910} & 25 \\
~~+NegTMR suite (ours) & \textbf{0.760} & 0.483 & \textbf{0.536} & \textbf{0.738} & 0.537 & \textbf{0.435} & \textbf{0.887} & 0.547 & \textbf{0.750} & 0.953 & 0.511 & 0.903 & 24 \\
\midrule
SnapMoGen evaluator & 0.965 & 0.518 & 0.927 & 0.728 & 0.519 & 0.435 & \textbf{0.838} & 0.529 & \textbf{0.656} & 0.998 & 0.500 & 0.995 & 1 \\
~~+NegSnap suite (ours) & \textbf{0.970} & 0.539 & \textbf{0.935} & \textbf{0.742} & 0.488 & \textbf{0.496} & 0.833 & 0.529 & 0.646 & \textbf{0.999} & 0.482 & \textbf{0.998} & 1 \\
\bottomrule
\end{tabular}
}
\end{center}
\end{table}

% 【翻译】定性失败案例。
\section{Qualitative failure cases}
\label{app:failure-gallery}
\label{app:examples}

% 【翻译】此处我们展示更多现有评估器扰动后失效的案例。动作固定，描述逐字
%   保留；展示为负的原始分差 sim(m,c)−sim(m,c')。每行十个均匀采样
%   的 SMPL-H 姿态按时间从左到右、从浅到深排列。

Here we provide more visualization cases where existing evaluators fail to detect the perturbations. Motions are fixed and captions are perturbed. Scores are negative raw
$\mathrm{sim}(m,c)-\mathrm{sim}(m,c')$ margins.
Each row shows ten uniformly sampled SMPL-H poses, earlier to later from left to right and light to dark.

\begin{figure}[!htbp]
\centering
\includegraphics[width=\linewidth]{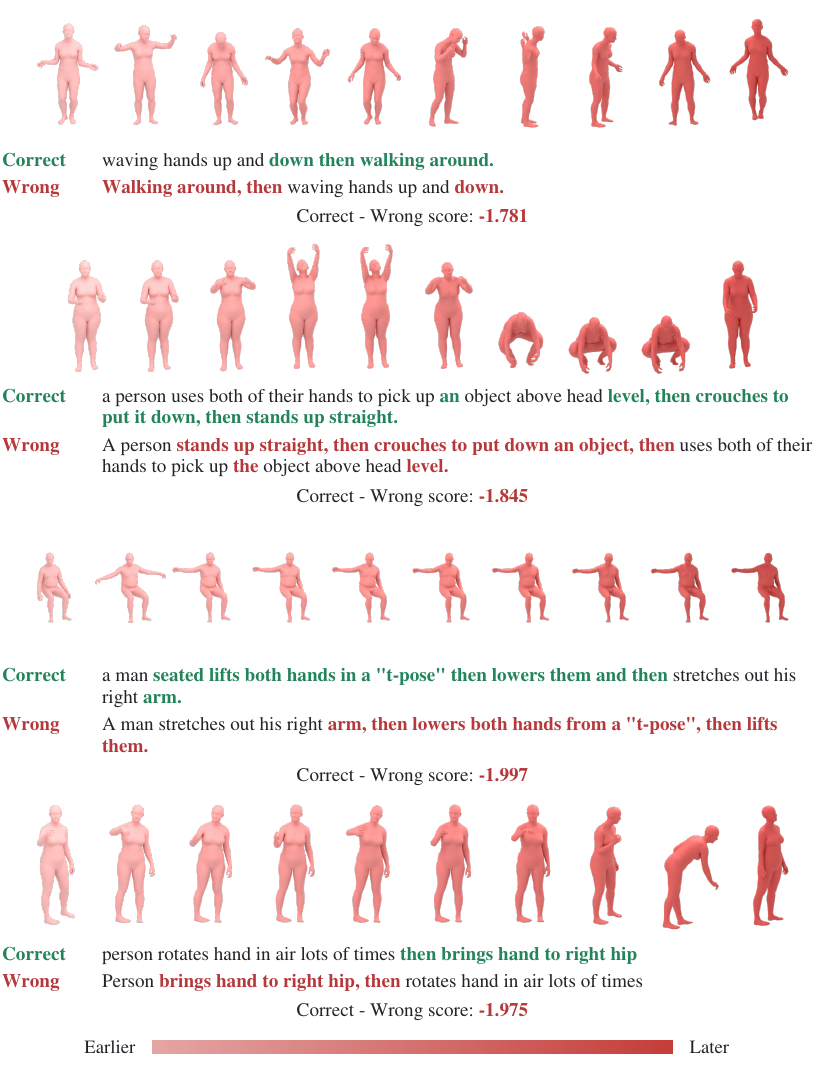}
% 【翻译】评测器的时间顺序失败案例。
\caption{Temporal Order failures of the evaluators.}
\label{fig:failure-order}
\end{figure}
\FloatBarrier
\clearpage

\begin{figure}[!htbp]
\centering
\includegraphics[width=\linewidth]{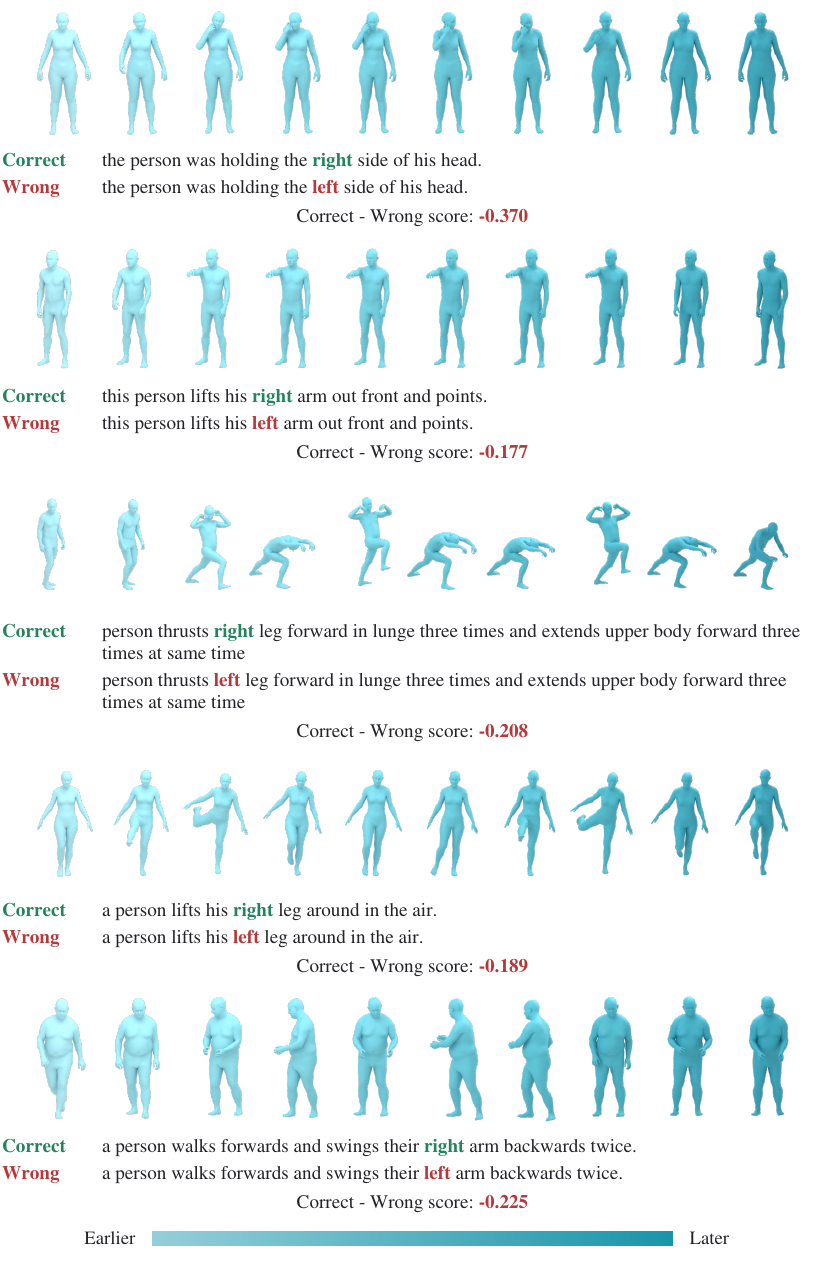}
% 【翻译】评测器的镜像反射失败案例。
\caption{Mirror Reflection failures of the evaluators.}
\label{fig:failure-mirror}
\end{figure}
\FloatBarrier
\clearpage

\begin{figure}[!htbp]
\centering
\includegraphics[width=\linewidth]{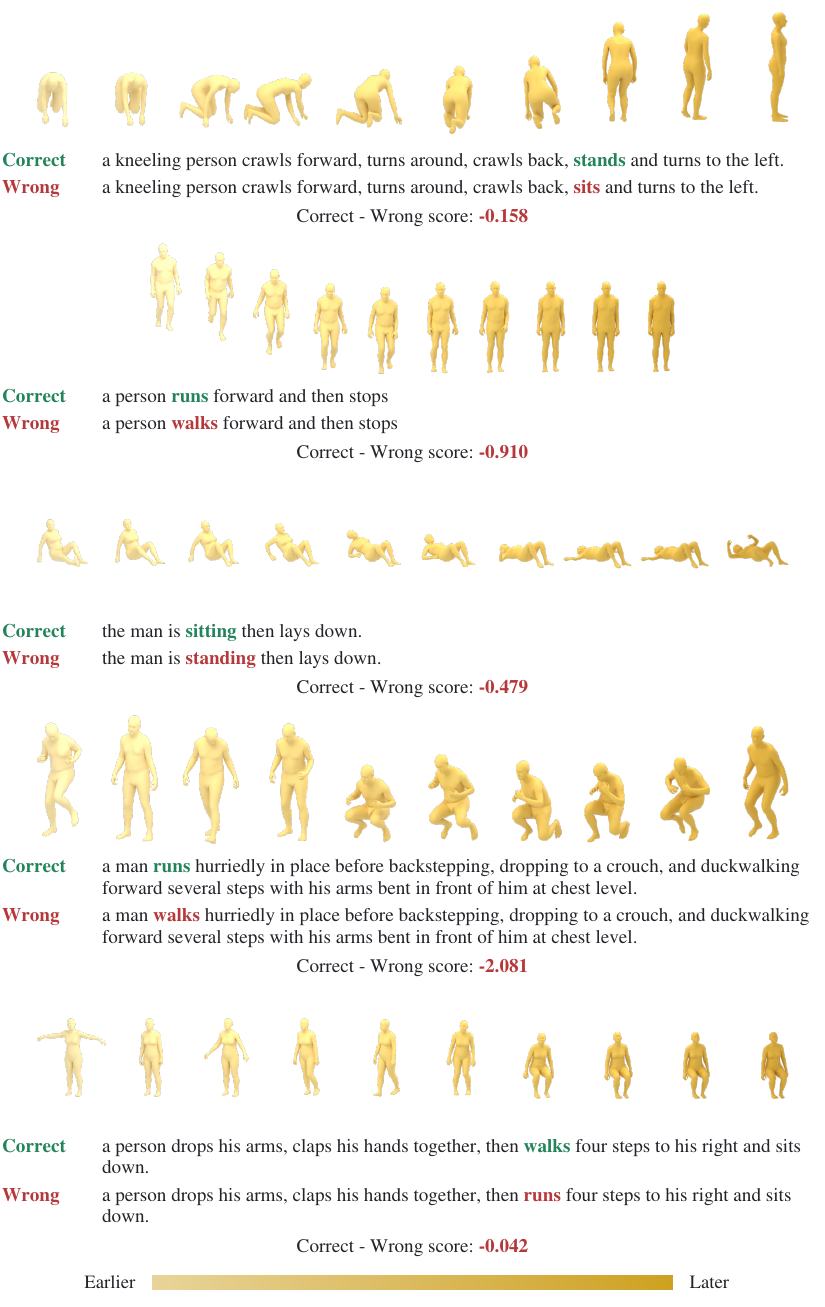}
% 【翻译】评测器的动作身份失败案例。
\caption{Action Identity failures of the evaluators.}
\label{fig:failure-action}
\end{figure}
\FloatBarrier
\end{document}